\documentclass[11pt,letterpaper,logo]{style}

\usepackage{amsmath,amsfonts,bm}

\def\eqref#1{equation~\ref{#1}}

\def\1{\bm{1}}

\DeclareMathAlphabet{\mathsfit}{\encodingdefault}{\sfdefault}{m}{sl}
\SetMathAlphabet{\mathsfit}{bold}{\encodingdefault}{\sfdefault}{bx}{n}

\usepackage{algorithm}
\usepackage{algpseudocode}
\usepackage{multirow}
\usepackage{fontawesome5}
\tcbuselibrary{skins,breakable}
\definecolor{promptaccent}{HTML}{537A93}
\graphicspath{{figures/}}
\newcommand{\CodeRepository}{https://github.com/ModalityDance/EngramEdit}

\newtcolorbox{propositionbox}{
    enhanced,
    breakable,
    colback=black!1.5!white,
    colframe=black!45!white,
    boxrule=0.45pt,
    arc=0.8mm,
    boxsep=0pt,
    left=8pt,
    right=8pt,
    top=6pt,
    bottom=6pt,
    before skip=8pt,
    after skip=0pt
}

\title{EngramEdit: Decoupled Knowledge Updates in LLMs through Conditional Memory}
\runningtitle{EngramEdit: Decoupled Knowledge Updates in LLMs through Conditional Memory}
\PublicDate{2026-10-08}
\author{%
    {\Authfont
    Hongru~Cai\textsuperscript{1}\quad
    Ran~Wei\textsuperscript{2}\quad
    Wenjie~Wang\textsuperscript{3}\quad
    Chengfa~Wu\textsuperscript{2}\authorcr
    Ning~Song\textsuperscript{2}\quad
    Yongqi~Li\textsuperscript{1}\advisor \quad
    Wenjie~Li\textsuperscript{1}}\\
    {\Affilfont
    \textsuperscript{1}The Hong Kong Polytechnic University \quad
    \textsuperscript{2}Hangzhou Diagens Biotechnology Co., Ltd\\
    \textsuperscript{3}University of Science and Technology of China}\\
    \texttt{\{henry.hongrucai, liyongqi0\}@gmail.com}
}

\begin{document}

\begin{abstract}
Conditional memory architectures such as DeepSeek Engram use input $n$-grams to look up learned embeddings, expanding the capacity of large language models (LLMs) with limited additional computation.
Beyond model scaling, this architecture has demonstrated the potential to decouple factual knowledge storage from general-purpose computation, offering a promising route to updating factual knowledge while keeping the Transformer backbone fixed.
Realizing this potential is challenging because different expressions of a fact may activate different $n$-gram embeddings, while updating shared embeddings can unintentionally change the model’s predictions about other facts.
We propose \textbf{EngramEdit} for decoupled knowledge updates through conditional memory. EngramEdit first computes target memory representations that make the model predict the updated fact across multiple expressions.
It then jointly updates the shared $n$-gram embeddings to match these targets across expressions and edits, penalizing updates to frequently reused embeddings more strongly to preserve unrelated knowledge.
Experiments show that EngramEdit enables independent factual knowledge updates through conditional memory, achieving near-perfect editing success.
Revised knowledge is usable across unseen expressions and in multi-hop reasoning, with nearly 3$\times$ the strongest baseline's accuracy under chain-of-thought (CoT) prompting.
Unrelated knowledge and general capabilities are largely preserved even as factual updates accumulate.
These findings show that EngramEdit turns conditional memory into an editable knowledge interface, extending its role beyond model scaling to support decoupled knowledge updates.
\end{abstract}

\newcommand{\TitleLinks}{%
\centering
    \vspace{6pt}
    {\noindent\absfont\fontsize{11}{13}\selectfont
    \faGithub\ GitHub: \url{https://github.com/ModalityDance/EngramEdit}\par}%
}

\maketitle

\section{Introduction}
\label{sec:introduction}

\begin{figure}[htbp]
    \centering
    \includegraphics[width=\linewidth]{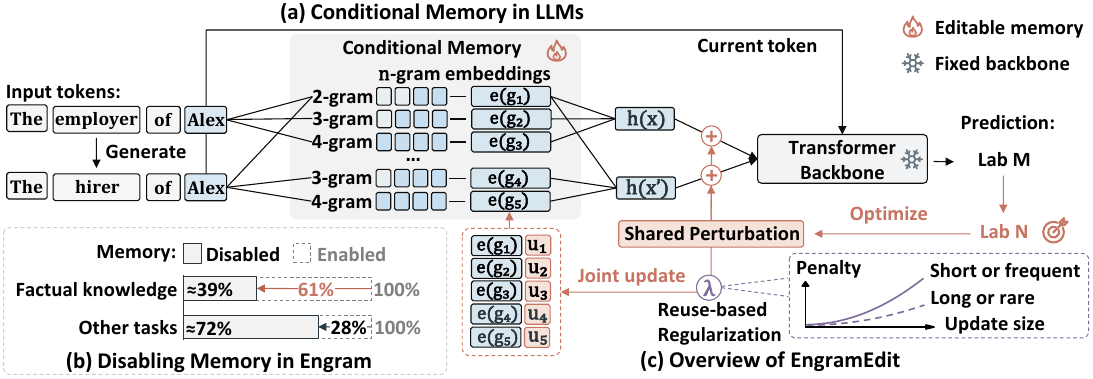}
    \caption{\textbf{Conditional memory and knowledge updating with EngramEdit}.
    (a) Input $n$-grams look up learned embeddings for backbone computation.
    (b) Disabling memory in DeepSeek Engram causes larger relative performance drops on factual-knowledge benchmarks~\citep{cheng2026engram}.
    (c) EngramEdit computes target memory representations and jointly updates the corresponding $n$-gram embeddings, with stronger penalties on frequently reused embeddings.}
    \label{fig:overview}
\end{figure}

Sparsity in large language models (LLMs) provides a promising way to scale model capacity without proportionally increasing per-token computation. Mixture-of-Experts (MoE)~\citep{shazeer2017moe,lepikhin2021gshard} achieves this through \emph{conditional computation}, activating only a subset of expert networks for each token. More recently, DeepSeek Engram~\citep{cheng2026engram} explores a complementary approach through \emph{conditional memory}. At each token position, the model uses $n$-grams of different lengths ending at that position to look up their learned embeddings. These embeddings provide input-dependent memory representations for the Transformer backbone computation (see Figure~\ref{fig:overview}(a)). This design is increasingly used to expand the capacity of LLMs such as DeepSeek-V4.1-Flash~\citep{deepseekai2026deepseekv41flash} and Qwen3.8-Flash-Next~\citep{qwen2026design}.

Beyond model scaling, conditional memory offers a promising architectural basis for decoupling factual knowledge storage from general-purpose computation in LLMs. In DeepSeek Engram, $n$-gram embeddings store reusable local representations, while the Transformer backbone dynamically integrates and processes these representations in context~\citep{cheng2026engram}. Ablation experiments on Engram further show that disabling conditional memory severely degrades factual knowledge performance while largely preserving the ability to understand and reason about information provided in context (see Figure~\ref{fig:overview}(b)).

This division of roles suggests a promising possibility: independently update factual knowledge in conditional memory without modifying the general-purpose backbone. Such updates would turn conditional memory into an editable knowledge interface, allowing LLMs to revise outdated facts while largely preserving general capabilities. Prior work mainly studies model scaling~\citep{cheng2026engram,liu2026scalingembeddings}, personalization~\citep{li2026userasengram}, and domain adaptation~\citep{hou2026engramadapter}, leaving this potential for decoupled knowledge updates largely unexplored.

However, updating factual knowledge through conditional memory is non-trivial because memory lookup is determined by $n$-grams rather than facts. 1) \textbf{\emph{Efficacy.}} Different facts being edited may require conflicting changes to a shared $n$-gram embedding, so an update that makes one edit succeed may cause another to fail. 2) \textbf{\emph{Generalization.}} Rephrasing the same fact may activate different $n$-grams whose embeddings were not updated, causing the edit to fail on the new expression. 3) \textbf{\emph{Specificity.}} Embeddings associated with shorter or more frequently used $n$-grams are also activated by many unrelated inputs, so updating them may change unrelated factual predictions.

To enable decoupled knowledge updates through conditional memory, we propose \textbf{EngramEdit}. The key idea is to first compute memory representations that make the LLM predict the updated facts, and then jointly translate these target representations into updates to the underlying shared $n$-gram embeddings. As shown in Figure~\ref{fig:overview}(c), we first generate multiple expressions of each fact to broaden $n$-gram coverage. For each edit, a shared learnable perturbation is temporarily added to the memory representations of its expressions. Optimizing only this perturbation by backpropagating the prediction loss for the new fact produces a target memory representation for each expression (see Sec.~\ref{sec:conditional_memory_target_computation}). We then map each target to the $n$-gram embeddings used by its expression, identifying embeddings shared across expressions and edits (see Sec.~\ref{sec:multi_expression_memory_mapping}). The embedding updates are solved jointly to match these targets, with each shared embedding receiving one update that accounts for all expressions and edits using it. To preserve unrelated knowledge, we include stronger penalties on large updates to frequently reused embeddings in this joint optimization (see Sec.~\ref{sec:joint_memory_updating}).

Our experiments show that EngramEdit enables effective and generalizable factual knowledge updates through conditional memory while largely preserving unrelated knowledge and general capabilities. 1) \textbf{Factual knowledge can be effectively updated.} By updating only conditional memory, EngramEdit achieves near-perfect editing success on both CounterFact~\citep{meng2022rome} and ZsRE~\citep{levy2017zsre}. 2) \textbf{Updated knowledge remains usable.} EngramEdit achieves strong generalization to unseen expressions, and the revised facts support multi-hop reasoning. On MQuAKE~\citep{zhong2023mquake}, it achieves nearly 3$\times$ the strongest baseline's multi-hop accuracy under chain-of-thought (CoT) prompting. 3) \textbf{Unrelated knowledge and general capabilities are largely preserved.} Overall performance on unrelated factual queries and general-ability tasks remains largely unchanged during extended sequential editing. Together, these findings show that EngramEdit turns conditional memory into an editable knowledge interface for decoupled knowledge updates. This extends its role beyond model scaling and opens a promising path toward LLMs that continually update their knowledge while largely preserving their general capabilities.

\section{Preliminaries}
\label{sec:preliminaries}

This section introduces factual knowledge editing (KE) and the structure of conditional memory.

\paragraph{Factual Knowledge Editing.} KE aims to update specific facts in a pretrained LLM $\mathcal{M}$ while preserving unrelated knowledge~\citep{meng2022rome,meng2023memit}. Each edit request $e_i=(s_i,r_i,o_i\rightarrow o_i^\star)$ replaces the original object $o_i$ with the desired object $o_i^\star$ for subject $s_i$ and relation $r_i$, e.g., changing Alex's employer from Lab M to Lab N. Given an edit prompt $x_i$ expressing $(s_i,r_i)$, the edited model should predict the desired object $o_i^\star$. In sequential editing, we process batches $\mathcal{E}=\{e_i\}_{i=1}^{m}$ of $m$ requests, applying each batch to the model updated by preceding batches.

EngramEdit shares a basic idea with locate-then-edit methods~\citep{meng2022rome,meng2023memit,fang2025alphaedit,gu2026moeedit}: first find a representation that makes the model predict the desired object, then update the parameters that produce it. We introduce this process through feed-forward network (FFN) editing. The editor first selects an FFN layer $\ell$ and a subject-token position $q_i$ associated with factual recall, typically the last subject token~\citep{meng2022rome}. At this location, the FFN output projection $\mathbf{W}_{\mathrm{out}}^\ell$ maps its input activation $\mathbf{k}_i^\ell$ to the current output $\mathbf{v}_i^\ell=\mathbf{W}_{\mathrm{out}}^\ell\mathbf{k}_i^\ell$. The editor temporarily adds a learnable perturbation $\boldsymbol{\delta}_i$ to this output and optimizes it by backpropagating the prediction loss for $o_i^\star$, with all model parameters fixed. The optimized perturbation $\boldsymbol{\delta}_i^\star$ gives the target output
\begin{equation}
    \mathbf{v}_i^\star
    =\mathbf{v}_i^\ell+\boldsymbol{\delta}_i^\star.
    \label{eq:target_ffn_output}
\end{equation}
The editor then solves for and applies a parameter update $\Delta\mathbf{W}_{\mathrm{out}}^\ell$ so that the FFN produces this target without the temporary perturbation (see App.~\ref{app:locate_then_edit} for the full formulation):
\begin{equation}
    (\mathbf{W}_{\mathrm{out}}^\ell+\Delta\mathbf{W}_{\mathrm{out}}^\ell)
    \mathbf{k}_i^\ell
    \approx\mathbf{v}_i^\star.
    \label{eq:locate_then_edit_parameter_update}
\end{equation}

\paragraph{Conditional Memory via $n$-gram Embeddings.} Conditional memory associates each $n$-gram $g$ with a learned embedding $\mathbf{e}(g)$ used when $g$ is activated~\citep{cheng2026engram,liu2026scalingembeddings,qwen2026design}. We treat each embedding as one logical vector, abstracting the finite hashed tables and sub-tables used to control storage costs in practical architectures~\citep{cheng2026engram,liu2026scalingembeddings}. The same $n$-gram uses the same embedding across inputs.

Given an input token sequence $x=(t_1,\ldots,t_T)$, let $g_{q,n}(x)=t_{q-n+1:q}$ denote the length-$n$ suffix $n$-gram ending at position $q$. For the configured length set $\mathcal{N}$, the activated $n$-gram set is $\mathcal{G}(x,q)=\{g_{q,n}(x):n\in\mathcal{N},\,n\leq q\}$. We write the contribution of the activated embeddings to the backbone computation as a memory representation $\mathbf{h}(x,q)$:
\begin{equation}
    \mathbf{h}(x,q)
    =
    \operatorname{Agg}_{x,q}
    \left(
        \{\mathbf{e}(g):g\in\mathcal{G}(x,q)\}
    \right),
    \label{eq:ngram_lookup}
\end{equation}
where $\operatorname{Agg}_{x,q}$ includes the architecture-specific processing of the activated embeddings and any interaction with the backbone state at position $q$. For example, additive aggregation combines the embeddings directly~\citep{liu2026scalingembeddings}, whereas context-aware gated fusion uses the backbone state to control their contribution~\citep{cheng2026engram}.

\section{Method}
\label{sec:method}

\begin{figure}[t]
    \centering
    \includegraphics[width=\linewidth]{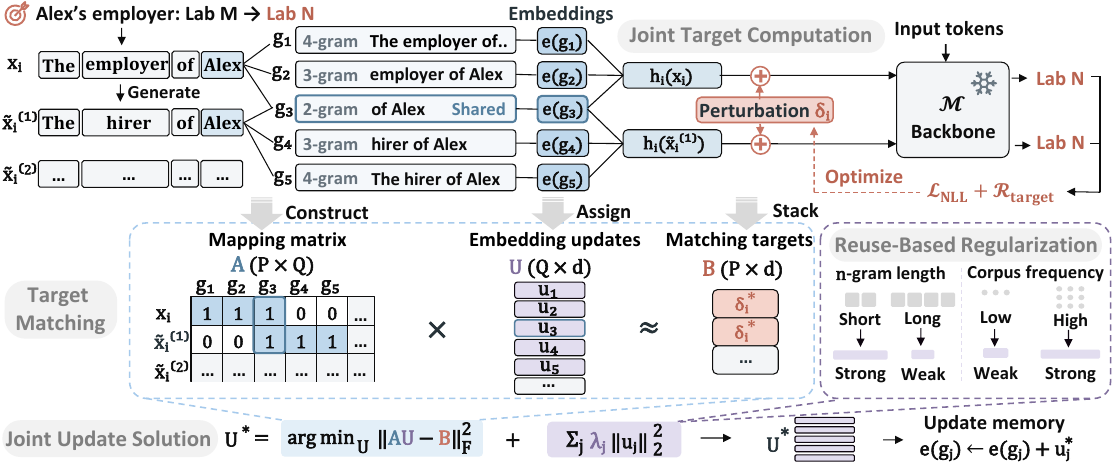}
    \caption{\textbf{Overview of EngramEdit}. We compute target memory representations that predict updated facts across expressions, map expressions to their activated $n$-grams, and jointly update the embeddings to match these targets with stronger penalties for shorter or more frequent \(n\)-grams.}
    \label{fig:method}
\end{figure}

In this section, we present EngramEdit for decoupled knowledge updates through conditional memory (see Figure~\ref{fig:method}). For each edit batch, we first compute target memory representations that make the LLM predict the updated facts across expressions (see Section~\ref{sec:conditional_memory_target_computation}). We then map each expression to its activated $n$-grams, identifying those shared across expressions and edits (see Section~\ref{sec:multi_expression_memory_mapping}). Finally, we jointly solve for a single update per embedding to match the targets using reuse-based regularization (see Section~\ref{sec:joint_memory_updating}). Algorithm~\ref{alg:engramedit} in App.~\ref{app:complete_editing_procedure} summarizes the complete procedure.

\subsection{Conditional Memory Target Computation}
\label{sec:conditional_memory_target_computation}

EngramEdit first learns how memory representations should change to predict the updated facts, providing targets for jointly allocating updates across shared \(n\)-gram embeddings.

\paragraph{Expression Generation.}
Different expressions of the same edit may activate different $n$-grams and use different embeddings. To cover more of these $n$-grams, we ask the model $\mathcal{M}$ to generate $K$ semantically equivalent expressions of the subject--relation request for each edit $e_i$ in the batch $\mathcal{E}$ before editing. The generation prompt encourages varied wording immediately before the subject (see App.~\ref{app:expression_generation} for detailed prompts). Combining the generated expression set $\mathcal{P}_i=\{\widetilde{x}_i^{(k)}\}_{k=1}^{K}$ with the original edit prompt $x_i$ gives $\mathcal{X}_i=\{x_i\}\cup\mathcal{P}_i$, used for target computation and memory mapping in Section~\ref{sec:multi_expression_memory_mapping}.

\paragraph{Joint Target Computation.}
We next compute target memory representations that make the LLM predict the updated fact across expressions. Since these expressions describe the same factual update, a shared perturbation allows them to jointly guide a single adjustment.

For each expression $x\in\mathcal{X}_i$, let $q_i(x)$ denote its last subject-token position~\citep{meng2022rome} and $\mathbf{h}_i(x)=\mathbf{h}(x,q_i(x))\in\mathbb{R}^{d}$ its memory representation there. Following~\cite{meng2022rome,meng2023memit}, we temporarily add a shared learnable perturbation $\boldsymbol{\delta}_i\in\mathbb{R}^{d}$ to each representation (see Figure~\ref{fig:method}). Let $\mathcal{M}_{\boldsymbol{\delta}_i}$ denote the model evaluated with this perturbation. With all model parameters fixed, we optimize only $\boldsymbol{\delta}_i$ using the prediction loss for the updated fact averaged across expressions:
\begin{equation}
\begin{aligned}
    \boldsymbol{\delta}_i^\star
    = \arg\min_{\boldsymbol{\delta}_i}\;&
    \frac{1}{|\mathcal{X}_i|}
    \sum_{x\in\mathcal{X}_i}
    \mathcal{L}_{\mathrm{NLL}}
    \bigl(\mathcal{M}_{\boldsymbol{\delta}_i}(x),o_i^\star\bigr)
    +\mathcal{R}_{\mathrm{target}}(\boldsymbol{\delta}_i),\\
    \text{subject to}\;&
    \|\boldsymbol{\delta}_i\|_2\leq\rho\|\mathbf{h}_i(x_i)\|_2.
\end{aligned}
\label{eq:target_optimization}
\end{equation}
Here, $\mathcal{L}_{\mathrm{NLL}}$ is the negative log-likelihood (NLL) loss for the updated fact. $\mathcal{R}_{\mathrm{target}}$ penalizes large perturbations and prediction changes when describing the subject~\citep{meng2022rome}, and the clamp factor $\rho>0$ bounds the perturbation norm (see App.~\ref{app:target_computation_objective} for details). The optimized perturbation $\boldsymbol{\delta}_i^\star$ defines a target memory representation for each expression:
\begin{equation}
    \mathbf{h}_i^\star(x)
    = \mathbf{h}_i(x)+\boldsymbol{\delta}_i^\star,
    \qquad x\in\mathcal{X}_i.
    \label{eq:conditional_memory_target}
\end{equation}

\subsection{Multi-Expression Memory Mapping}
\label{sec:multi_expression_memory_mapping}

EngramEdit next links each target memory representation to the $n$-gram embeddings activated by its expression. Because expressions and edits may share embeddings, we construct a single mapping for the entire edit batch.

\paragraph{Memory Mapping.}
To link each target memory representation to its underlying embeddings, we identify the $n$-grams activated at the expression's last subject-token position. For each expression $x\in\mathcal{X}_i$, the activated $n$-gram set is $\mathcal{G}_i(x)=\mathcal{G}(x,q_i(x))$, using the lookup notation from Section~\ref{sec:preliminaries}. The union of these sets, $\mathcal{G}_i=\bigcup_{x\in\mathcal{X}_i}\mathcal{G}_i(x)$, collects the activated $n$-grams for edit $e_i$. Expressions activating the same $n$-gram $g$ share its embedding $\mathbf{e}(g)$, as illustrated by ``of Alex'' (see Figure~\ref{fig:method}). When combining the $n$-gram sets $\mathcal{G}_i$ across the batch, we therefore retain each $n$-gram only once.

\paragraph{Mapping Matrix.}
To jointly allocate updates to shared embeddings, we record which expressions use each embedding in a batch-level mapping matrix $\mathbf{A}\in\mathbb{R}^{P\times Q}$. Each of the $P=\sum_{i=1}^{m}|\mathcal{X}_i|$ rows corresponds to one expression and its target memory representation. The $Q$ columns correspond to the distinct $n$-grams $\{g_j\}_{j=1}^{Q}$ in $\bigcup_{i=1}^{m}\mathcal{G}_i$, with column $j$ representing the embedding $\mathbf{e}(g_j)$. Row $p$ is indexed by an edit--expression pair $(i_p,x_p)$, where the expression $x_p\in\mathcal{X}_{i_p}$. The matrix element $A_{pj}$ is 1 if expression $x_p$ activates $n$-gram $g_j$, and 0 otherwise:
\begin{equation}
    A_{pj}
    = \mathbb{I}\!\left[g_j\in\mathcal{G}_{i_p}(x_p)\right].
    \label{eq:memory_mapping_matrix}
\end{equation}

\subsection{Joint Memory Update Allocation}
\label{sec:joint_memory_updating}

Using the expression-to-$n$-gram mapping, EngramEdit jointly computes embedding updates to match targets while penalizing large updates to frequently reused embeddings.

\paragraph{Target Matching.}
To match each target memory representation, we express its required change in terms of updates to the activated embeddings. We derive this relation under sum aggregation, where a memory representation changes by the sum of its embedding updates~\citep{liu2026scalingembeddings}. For column $j$ of the mapping matrix $\mathbf{A}$, let $\mathbf{u}_j\in\mathbb{R}^{d}$ denote the update to the $n$-gram embedding $\mathbf{e}(g_j)$.

For row $p$ of the mapping matrix $\mathbf{A}$, let $\mathbf{h}_p=\mathbf{h}_{i_p}(x_p)$ denote the current memory representation and $\mathbf{h}_p^\star=\mathbf{h}_{i_p}^\star(x_p)$ its target. To match this target, the combined embedding updates must approximate the difference between the two representations:
\begin{equation}
    \sum_{j=1}^{Q}A_{pj}\mathbf{u}_j
    \approx
    \mathbf{h}_p^\star-\mathbf{h}_p
    =\boldsymbol{\delta}_{i_p}^\star.
    \label{eq:memory_representation_matching}
\end{equation}
We denote this difference by the matching target $\mathbf{b}_p=\mathbf{h}_p^\star-\mathbf{h}_p$. Stacking the embedding updates as rows gives the embedding update matrix $\mathbf{U}=[\mathbf{u}_1,\ldots,\mathbf{u}_Q]^\top\in\mathbb{R}^{Q\times d}$; stacking the matching targets gives the matching target matrix $\mathbf{B}\in\mathbb{R}^{P\times d}$. The batch-level matching requirement is therefore $\mathbf{A}\mathbf{U}\approx\mathbf{B}$ (see Figure~\ref{fig:method}). App.~\ref{app:gated_aggregation} extends this formulation to context-aware gated aggregation used in Engram~\citep{cheng2026engram}.

\paragraph{Reuse-Based Regularization.}
To preserve unrelated knowledge, we penalize large updates more strongly for embeddings likely to be activated by unrelated inputs. We use $n$-gram length and corpus frequency as indicators of activation frequency across inputs, assigning larger regularization weights $w_j$ to shorter or more frequent $n$-grams (see App.~\ref{app:protection_weight} for the construction of the weights). The regularization matrix is
\begin{equation}
    \boldsymbol{\Lambda}
    =\operatorname{diag}(\lambda_1,\ldots,\lambda_Q),
    \qquad
    \lambda_j
    =\lambda_{\mathrm{reuse}}w_j+\lambda_{\mathrm{ridge}},
    \label{eq:reuse_regularizer}
\end{equation}
where $\lambda_{\mathrm{reuse}}$ controls the reuse-based penalty, while the optional ridge term $\lambda_{\mathrm{ridge}}$ independently sets a uniform penalty on all updates. Both coefficients are nonnegative, with $\lambda_j>0$ for every embedding. We prove that, under sum aggregation, weighting update penalties by activation probability bounds the average squared change in memory representations on unrelated inputs, providing a theoretical basis for reuse-based regularization (see App.~\ref{app:reuse_protection} for the proof and bounds on individual embedding updates).

\paragraph{Joint Update Solution.}
We combine target matching and reuse-based regularization to solve for the embedding updates:
\begin{equation}
    \mathbf{U}^\star
    =\arg\min_{\mathbf{U}}
    \left\|\mathbf{A}\mathbf{U}-\mathbf{B}\right\|_F^2
    +\sum_{j=1}^{Q}\lambda_j\left\|\mathbf{u}_j\right\|_2^2.
    \label{eq:joint_memory_update}
\end{equation}
By jointly considering all expressions that share an embedding, this objective balances their potentially conflicting targets (see App.~\ref{app:matching_structure} for exact-matching conditions). With all $\lambda_j>0$, $\mathbf{A}^\top\mathbf{A}+\boldsymbol{\Lambda}$ is positive definite, yielding a unique update computed by solving the corresponding linear system (see App.~\ref{app:joint_update_solution_proof} for the proof):
\begin{equation}
    \mathbf{U}^\star
    =
    \left(\mathbf{A}^\top\mathbf{A}+\boldsymbol{\Lambda}\right)^{-1}
    \mathbf{A}^\top\mathbf{B}.
    \label{eq:joint_memory_update_solution}
\end{equation}
We prove that the joint solution has two useful properties. 1) \textbf{Minimum update cost.} Different embedding updates can produce the same memory representation changes but incur different update costs. EngramEdit minimizes the weighted embedding update cost among all updates producing the same achieved memory representation changes $\mathbf{A}\mathbf{U}^\star$ (see App.~\ref{app:minimum_cost_allocation} for the proof). 2) \textbf{Stability to target changes.} Since targets are computed through optimization, they may vary slightly. With the mapping and regularization fixed, we prove that regularization bounds the extent to which these variations affect the solved embedding updates (see App.~\ref{app:solution_sensitivity} for the analysis and proof).

\paragraph{Memory Updating.}
Finally, we update conditional memory as $\mathbf{e}(g_j)\leftarrow\mathbf{e}(g_j)+\mathbf{u}_j^\star$, keeping all other parameters fixed. During inference, an embedding update enters the model's computation only when its $n$-gram is active in the current prefix.

\section{Experiments}
\label{sec:experiments}

In this section, we conduct experiments to address the following research questions:
\begin{itemize}[leftmargin=1em, labelsep=0.4em, nosep]
    \item \textbf{RQ1:} How effectively does EngramEdit update factual knowledge through conditional memory while preserving unrelated knowledge?
    \item \textbf{RQ2:} How well does EngramEdit preserve general capabilities during knowledge updates?
    \item \textbf{RQ3:} How do EngramEdit's design choices affect editing performance?
    \item \textbf{RQ4:} How do EngramEdit's memory updates support the LLM's use of revised knowledge?
\end{itemize}

\subsection{Experimental Setup}
\label{sec:experimental_setup}

\paragraph{Datasets and Metrics.}
On CounterFact~\citep{meng2022rome} and ZsRE~\citep{levy2017zsre}, we report Efficacy (success on edited prompts), Generalization (success on held-out paraphrases not used during editing), Specificity (performance on unrelated queries), and their arithmetic mean, Utility. On CounterFact, we also report Fluency and Consistency for generated text. MQuAKE~\citep{zhong2023mquake} measures multi-hop use of updated facts through answer accuracy. Following AlphaEdit~\citep{fang2025alphaedit}, we report mean task-level F1 across six general-ability tasks. See Apps.~\ref{app:datasets} and~\ref{app:evaluation_metrics} for dataset examples and metric definitions.

\paragraph{Baselines and Implementation.}
We compare editors applicable to MoE LLMs: FT, FT-L~\citep{zhu2020modifying}, AdaLoRA~\citep{zhang2023adalora}, UnKE~\citep{deng2025unke}, and MoEEdit~\citep{gu2026moeedit}. To compare direct memory fine-tuning, we include Memory-FT (Subject) and Memory-FT (All Tokens), which update $n$-gram embeddings activated at the last subject token or all token positions in the original edit prompts, respectively. All experiments use LongCat-Flash-Lite~\citep{liu2026scalingembeddings}, with sequential batch editing. See Apps.~\ref{app:baseline_methods} and~\ref{app:implementation_details} for details.

\begin{table}[t]
    \centering
    \caption{\textbf{Editing results on CounterFact and ZsRE}. MFT-S and MFT-A denote Memory-FT (Subject) and (All Tokens). Eff. = Efficacy, Gen. = Generalization, Spec. = Specificity, Util. = Utility, Flu. = Fluency, Cons. = Consistency. Subscripts give 95\% confidence-interval half-widths. Bold and underlining mark the best and second-best editing scores.}
    \label{tab:main_editing_results}
    \begingroup
    \fontsize{8}{10}\selectfont
    \setlength{\tabcolsep}{1.4pt}
    \renewcommand{\arraystretch}{0.9}
    \newcommand{\tabci}[2]{\ensuremath{\text{#1}{}_{\scriptscriptstyle\pm\text{#2}}}}
    \newcommand{\metrichead}[1]{\textbf{#1}\,$\boldsymbol{\uparrow}$}
    \resizebox{\linewidth}{!}{%
    \begin{tabular}{@{}lcccccccccc@{}}
        \toprule
        \multirow{2}{*}[-0.5ex]{\textbf{Method}} & \multicolumn{6}{c}{\textbf{CounterFact}} & \multicolumn{4}{c}{\textbf{ZsRE}} \\
        \cmidrule(lr){2-7}\cmidrule(lr){8-11}
        & \metrichead{Eff.} & \metrichead{Gen.} & \metrichead{Spec.} & \metrichead{Util.} & \metrichead{Flu.} & \metrichead{Cons.} & \metrichead{Eff.} & \metrichead{Gen.} & \metrichead{Spec.} & \metrichead{Util.} \\
        \midrule
        Pre-edited & \tabci{9.5}{1.28} & \tabci{11.2}{1.19} & \tabci{86.8}{0.94} & 35.8 & \tabci{544.0}{1.67} & \tabci{15.0}{0.37} & \tabci{36.9}{1.30} & \tabci{36.5}{1.29} & \tabci{41.6}{1.19} & 38.3 \\
        \midrule
        FT & \tabci{94.9}{0.96} & \tabci{63.6}{1.83} & \tabci{43.8}{1.63} & 67.4 & \tabci{505.1}{2.04} & \tabci{13.3}{0.36} & \tabci{76.7}{1.13} & \tabci{76.8}{1.15} & \tabci{43.1}{1.27} & 65.5 \\
        FT-L & \tabci{15.3}{1.58} & \tabci{12.5}{1.24} & \tabci{\underline{83.3}}{1.11} & 37.0 & \tabci{542.0}{1.70} & \tabci{14.7}{0.36} & \tabci{58.0}{1.47} & \tabci{54.4}{1.48} & \tabci{45.8}{1.24} & 52.7 \\
        AdaLoRA & \tabci{61.1}{2.13} & \tabci{27.5}{1.55} & \tabci{51.1}{1.35} & 46.6 & \tabci{544.3}{1.57} & \tabci{12.9}{0.35} & \tabci{70.9}{1.40} & \tabci{64.7}{1.46} & \tabci{\underline{46.8}}{1.27} & 60.8 \\
        UnKE & \tabci{84.5}{1.59} & \tabci{58.0}{1.88} & \tabci{47.5}{1.61} & 63.3 & \tabci{546.5}{1.64} & \tabci{15.0}{0.35} & \tabci{\underline{86.9}}{1.13} & \tabci{\underline{79.0}}{1.33} & \tabci{\textbf{49.9}}{1.32} & \underline{71.9} \\
        MoEEdit & \tabci{\underline{99.1}}{0.41} & \tabci{\underline{67.1}}{1.69} & \tabci{63.5}{1.28} & 76.6 & \tabci{544.7}{1.65} & \tabci{14.7}{0.33} & \tabci{85.1}{0.91} & \tabci{78.7}{1.21} & \tabci{44.4}{1.26} & 69.4 \\
        MFT-S & \tabci{99.0}{0.45} & \tabci{59.1}{1.92} & \tabci{\textbf{85.2}}{0.90} & \underline{81.1} & \tabci{\underline{555.2}}{1.38} & \tabci{\underline{15.3}}{0.36} & \tabci{66.9}{1.35} & \tabci{56.5}{1.53} & \tabci{36.9}{1.17} & 53.4 \\
        MFT-A & \tabci{93.9}{1.04} & \tabci{40.7}{1.75} & \tabci{66.6}{1.12} & 67.1 & \tabci{496.7}{4.04} & \tabci{11.5}{0.40} & \tabci{62.8}{1.49} & \tabci{51.3}{1.57} & \tabci{33.7}{1.13} & 49.3 \\
        EngramEdit & \tabci{\textbf{99.5}}{0.31} & \tabci{\textbf{97.0}}{0.69} & \tabci{\textbf{85.2}}{0.92} & \textbf{93.9} & \tabci{\textbf{565.2}}{1.10} & \tabci{\textbf{16.1}}{0.36} & \tabci{\textbf{97.3}}{0.40} & \tabci{\textbf{93.7}}{0.79} & \tabci{38.3}{1.18} & \textbf{76.4} \\
        \bottomrule
    \end{tabular}%
    }
    \endgroup
\end{table}

\begin{figure}[t]
    \centering
    \includegraphics[width=\linewidth]{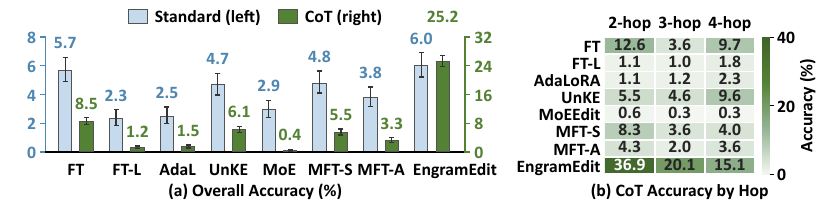}
    \caption{\textbf{Editing results on MQuAKE}. (a) Standard (left) and CoT (right) prompting with 95\% confidence intervals. (b) CoT accuracy by hop. AdaL and MoE denote AdaLoRA and MoEEdit.}
    \label{fig:mquake_overall}
\end{figure}

\subsection{Knowledge Update and Preservation (RQ1)}
\label{sec:overall_performance}

To evaluate how effectively EngramEdit updates facts through conditional memory, we compare it with the baselines after 2,000 sequential edits on each of CounterFact and ZsRE (see Table~\ref{tab:main_editing_results}). We further apply edits from 3,000 MQuAKE cases to test whether the updated facts support multi-hop reasoning (see Figure~\ref{fig:mquake_overall}). We apply standard prompting for direct answers and CoT prompting for step-by-step reasoning before answering. We observe that:

\begin{itemize}[leftmargin=1em, labelsep=0.4em, nosep]

\item On CounterFact, EngramEdit achieves similar Efficacy to MoEEdit and MFT-S but much higher Generalization; on ZsRE, it leads in both metrics (see Table~\ref{tab:main_editing_results}). Comparable success on edited prompts, therefore, does not imply comparable recall through other expressions. EngramEdit addresses this gap by updating embeddings across expressions, since different wordings can activate different $n$-grams. Its advantages persist through 5,000 sequential edits, showing that revised facts remain accessible as further updates accumulate (see App.~\ref{app:extended_editing_results} for details).

\item On MQuAKE, EngramEdit, and FT have similar accuracy under standard prompting. With CoT, EngramEdit achieves nearly 3$\times$ the accuracy of the strongest baseline and leads across all hop groups, as shown in Figure~\ref{fig:mquake_overall}. This improvement shows that facts updated through conditional memory can be used in multi-hop reasoning. During CoT reasoning, intermediate steps can activate updated embeddings for the facts used, allowing revised information to guide subsequent steps. The advantage extends through four-hop cases, supporting the use of updated knowledge in longer reasoning chains (see App.~\ref{app:mquake_hop_results} for full results).

\item On ZsRE, EngramEdit's Specificity remains close to the pre-edit level, whereas FT-L, AdaLoRA, UnKE, and MoEEdit exceed it. These higher scores do not by themselves establish better preservation, because Specificity counts newly correct answers as well as retained correct answers. The correctness transition analysis shows that corrections to previously wrong, unrelated predictions contribute to the baselines' gains and that EngramEdit retains most initially correct predictions (see App.~\ref{app:zsre_correctness_transitions} for further analysis). This provides additional evidence that its knowledge updates largely preserve performance on unrelated queries.

\item EngramEdit achieves the highest Utility on CounterFact and ZsRE, with Specificity close to pre-edit levels, and also leads in Fluency and Consistency. Together, these results show that EngramEdit supports effective and generalizable knowledge updates through conditional memory while largely preserving unrelated knowledge.

\end{itemize}

\subsection{General Capabilities (RQ2)}
\label{sec:general_capabilities}

To assess whether EngramEdit preserves general capabilities as knowledge updates accumulate, we evaluate six general-ability tasks in 5,000 sequential CounterFact edits. We compare mean task-level F1 and per-task changes against pre-edit performance. The results in Figure~\ref{fig:general_capabilities} show that:

\begin{figure}[t]
    \centering
    \includegraphics[width=\linewidth]{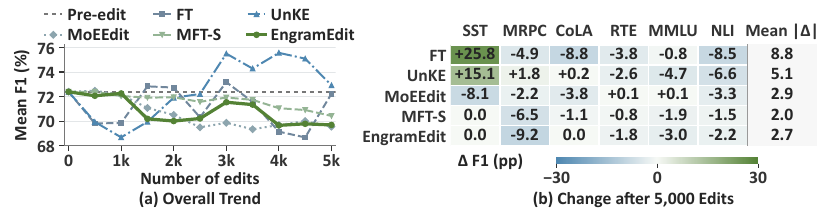}
    \caption{\textbf{General capabilities during sequential CounterFact editing}. (a) Mean F1 across six tasks in the editing sequence. (b) Task-level F1 changes from pre-edit scores after 5,000 edits, in percentage points (pp). The mean $|\Delta|$ averages the absolute task-level changes.}
    \label{fig:general_capabilities}
\end{figure}

\begin{itemize}[leftmargin=1em, labelsep=0.4em, nosep]
\item EngramEdit retains over 96\% of its pre-edit mean F1 after 5,000 sequential edits, with modest changes throughout the editing sequence, as shown in Figure~\ref{fig:general_capabilities}(a). MFT-S also largely retains pre-edit mean F1, but EngramEdit achieves much stronger editing Generalization (see Section~\ref{sec:overall_performance}). Both methods restrict editing to selected $n$-gram embeddings, and EngramEdit further penalizes large changes to frequently reused embeddings to limit effects on unrelated tasks. Together, these results support conditional memory as an editable knowledge interface and show that EngramEdit combines stronger cross-expression recall with largely preserved general capabilities.
\item EngramEdit shows small F1 changes on most tasks, with a larger decline on MRPC (see Figure~\ref{fig:general_capabilities}(b) and full trajectories in App.~\ref{app:task_level_capabilities}). MRPC tests whether two sentences express the same meaning. These sentences can activate different $n$-grams, so local embedding updates may affect their representations differently, altering the equivalence judgment.
\end{itemize}

\subsection{In-Depth Analysis}
\label{sec:in_depth_analysis}

\paragraph{Ablation Study (RQ3).}
\label{sec:ablation_studies}
To isolate each component's contribution, we compare five variants on CounterFact and ZsRE, with other settings held fixed. \emph{w/o Joint} solves embedding updates independently for each expression. \emph{w/o Freq.} and \emph{w/o Len.} remove the weighting that increases penalties for more frequent and shorter $n$-grams, respectively. \emph{w/o Reg.} removes reuse-based regularization, and \emph{w/o Reg. \& Expr.} additionally removes generated expressions. From Table~\ref{tab:reuse_regularization_ablation}, we observe that: 1) Removing joint allocation substantially lowers Efficacy and Generalization on both datasets. This shows that jointly considering the targets of all expressions sharing an embedding helps the model recall revised facts across expressions. 2) Removing generated expressions from the unregularized variant substantially lowers Generalization. Generated expressions, therefore, mainly support cross-expression recall by covering the different $n$-grams activated by alternative wordings. 3) Removing reuse-based regularization lowers Efficacy and Specificity, and removing either frequency or length weighting also reduces Utility. The regularization thus helps preserve unrelated knowledge by limiting changes to embeddings reused by other inputs. Our analysis further shows that over 92\% of new errors on unrelated queries involve activation of updated embeddings (see App.~\ref{app:trigger_attribution}).

\begin{table}[t]
    \centering
    \caption{\textbf{Ablations} of joint update allocation (Joint), reuse-based regularization (Reg.), $n$-gram corpus frequency (Freq.) and length (Len.) weighting, and generated expressions (Expr.).}
    \label{tab:reuse_regularization_ablation}
    \begingroup
    \fontsize{8}{10}\selectfont
    \setlength{\tabcolsep}{1pt}
    \renewcommand{\arraystretch}{1}
    \newcommand{\tabci}[2]{\ensuremath{\text{#1}{}_{\scriptscriptstyle\pm\text{#2}}}}
    \newcommand{\metrichead}[1]{\textbf{#1}\,$\boldsymbol{\uparrow}$}
    \begin{tabular*}{\linewidth}{@{\extracolsep{\fill}}lcccccccccc@{}}
        \toprule
        \multirow{2}{*}[-0.5ex]{\textbf{Variant}} & \multicolumn{6}{c}{\textbf{CounterFact}} & \multicolumn{4}{c}{\textbf{ZsRE}} \\
        \cmidrule(lr){2-7}\cmidrule(lr){8-11}
        & \metrichead{Eff.} & \metrichead{Gen.} & \metrichead{Spec.} & \metrichead{Util.} & \metrichead{Flu.} & \metrichead{Cons.} & \metrichead{Eff.} & \metrichead{Gen.} & \metrichead{Spec.} & \metrichead{Util.} \\
        \midrule
        EngramEdit & \tabci{\textbf{99.5}}{0.31} & \tabci{\textbf{97.0}}{0.69} & \tabci{\textbf{85.2}}{0.92} & \textbf{93.9} & \tabci{565.2}{1.10} & \tabci{\underline{16.1}}{0.36} & \tabci{\textbf{97.3}}{0.40} & \tabci{\underline{93.7}}{0.79} & \tabci{\textbf{38.3}}{1.18} & \textbf{76.4} \\
        w/o Joint & \tabci{74.5}{1.91} & \tabci{72.8}{1.74} & \tabci{84.8}{0.93} & 77.4 & \tabci{\textbf{568.0}}{1.00} & \tabci{15.3}{0.35} & \tabci{57.4}{1.55} & \tabci{58.3}{1.57} & \tabci{35.6}{1.17} & 50.4 \\
        w/o Freq. & \tabci{\underline{99.4}}{0.32} & \tabci{\underline{96.8}}{0.69} & \tabci{\textbf{85.2}}{0.92} & \underline{93.8} & \tabci{564.9}{1.22} & \tabci{\underline{16.1}}{0.36} & \tabci{\underline{96.0}}{0.52} & \tabci{92.6}{0.84} & \tabci{\underline{37.3}}{1.19} & 75.3 \\
        w/o Len. & \tabci{99.3}{0.38} & \tabci{\underline{96.8}}{0.70} & \tabci{\underline{85.1}}{0.93} & 93.7 & \tabci{565.4}{1.09} & \tabci{\textbf{16.2}}{0.36} & \tabci{95.7}{0.55} & \tabci{93.4}{0.75} & \tabci{37.2}{1.17} & 75.4 \\
        w/o Reg. & \tabci{98.8}{0.47} & \tabci{\underline{96.8}}{0.72} & \tabci{84.6}{0.93} & 93.4 & \tabci{\textbf{568.0}}{1.02} & \tabci{15.5}{0.35} & \tabci{\underline{96.0}}{0.54} & \tabci{\textbf{94.0}}{0.75} & \tabci{37.2}{1.19} & \underline{75.7} \\
        w/o Reg. \& Expr. & \tabci{98.8}{0.47} & \tabci{81.5}{1.59} & \tabci{\textbf{85.2}}{0.92} & 88.5 & \tabci{\underline{565.9}}{1.08} & \tabci{15.6}{0.36} & \tabci{93.9}{0.72} & \tabci{85.4}{1.19} & \tabci{37.0}{1.17} & 72.1 \\
        \bottomrule
    \end{tabular*}
    \endgroup
\end{table}

\begin{figure}[t]
    \centering
    \includegraphics[width=\linewidth]{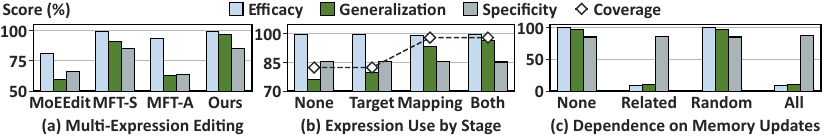}
    \caption{\textbf{Generated expressions and memory use on CounterFact}. (a) Editors use the same generated expressions per fact. (b) Generated expressions are used for target computation, memory mapping, both, or neither. Coverage is the percentage of held-out paraphrases activating fact-related updated embeddings. (c) Disabling no updates (None), fact-related updates (Related), matched random updates (Random), or all updates (All) during inference. }
    \label{fig:in_depth_analysis}
\end{figure}

\paragraph{Analysis of Generated Expressions (RQ3).}
\label{sec:expression_coverage}
To distinguish the benefit of additional expressions from how they are used, we first compare MoEEdit, MFT-S, MFT-A, and EngramEdit with the same generated expressions per fact on CounterFact (see Figure~\ref{fig:in_depth_analysis}(a)). We then vary expression use in EngramEdit's target computation and memory mapping, keeping other settings fixed (see Figure~\ref{fig:in_depth_analysis}(b)). All variants retain the original edit prompt. Coverage is the percentage of held-out paraphrases that activate an embedding updated for the same fact. We observe that: 1) With the same generated expressions, EngramEdit outperforms MFT-S in Generalization at similar Efficacy and Specificity. EngramEdit's target computation and joint memory updating thus use these expressions more effectively for recall under different wordings (see App.~\ref{app:multi_expression_editing} for full results). 2) Using generated expressions for memory mapping increases Coverage and yields most of the Generalization gain, because the expanded set of updated $n$-grams allows more held-out paraphrases to activate fact-related embeddings. An additional analysis of memory activation supports this explanation: the roughly 2\% of held-out paraphrases activating no fact-related embeddings account for half of generalization failures (see App.~\ref{app:trigger_attribution}). With Coverage unchanged, also using these expressions for target computation further improves Generalization by optimizing targets across expressions.

\paragraph{Analysis of Memory Use (RQ4).}
\label{sec:memory_update_analysis}

To test whether the LLM uses the updated conditional memory to predict revised facts, we compare normal inference with disabling updates for the evaluated fact, randomly selecting other updates matched in number and $n$-gram length, or all updates. All conditions use the same model after CounterFact editing and retain the original embeddings. As shown in Figure~\ref{fig:in_depth_analysis}(c), fact-related disabling sharply lowers Efficacy and Generalization, almost matching the effect of disabling all updates. In contrast, matched random disabling leaves Efficacy and Generalization unchanged. An additional analysis of prediction preferences shows that fact-related disabling makes the model favor original over revised facts on average (see App.~\ref{app:memory_disabling}). These results show that the LLM relies on the fact-related memory updates to access and utilize revised knowledge.

\paragraph{Additional Analyses.}
We extend editing to 5,000 facts to test stability (see App.~\ref{app:extended_editing_results}) and analyze ZsRE preservation (see App.~\ref{app:zsre_correctness_transitions}). We vary the number of generated expressions, editable $n$-gram lengths, regularization coefficients, and the clamp factor (see Apps.~\ref{app:expression_count}--\ref{app:regularization_clamp}). We further analyze memory use through memory activation and cross-edit sharing (see Apps.~\ref{app:memory_disabling}--\ref{app:cross_edit_sharing}).

\section{Related Work}
\label{sec:related-work}

\paragraph{Conditional Memory and $n$-gram Embeddings.}
Lookup-based memory architectures expand model capacity through sparse access to stored vectors, as in Product-Key Memory~\citep{lample2019productkeys} and Memory Layers at Scale~\citep{berges2025memorylayers}. A related line uses $n$-gram embeddings to associate local token or byte sequences with reusable representations.
1) \emph{Model scaling.} Byte Latent Transformer~\citep{pagnoni2025blt}, OverEncoding~\citep{huang2025overtokenized}, and SCONE~\citep{yu2025scone} use $n$-gram representations to enrich model inputs with local context at limited inference cost. Engram~\citep{cheng2026engram} and LongCat-Flash-Lite~\citep{liu2026scalingembeddings} further explore $n$-gram-based memory to scale LLM capacity. For memory construction, Memory Grafting~\citep{cheng2026memorygrafting} builds frozen conditional memory offline from another model's hidden states for pre-training. Related $n$-gram-based designs are also used in Qwen3.8-Flash-Next~\citep{qwen2026design} and DeepSeek-V4.1-Flash~\citep{deepseekai2026deepseekv41flash}.
2) \emph{Model adaptation.} Conditional memory has also been adapted for personalization through user-specific embedding updates in User as Engram~\citep{li2026userasengram} and for domain adaptation in Engram Adapter~\citep{hou2026engramadapter}. Our work develops EngramEdit to use pretrained conditional memory as an editable knowledge interface, enabling decoupled factual updates while largely preserving unrelated knowledge and general capabilities.

\paragraph{Knowledge Editing.}
KE methods can be grouped according to whether they modify the LLM's original parameters.
1) \emph{Parameter-preserving methods.} These methods supply updated knowledge through external information or additional components, keeping the original parameters fixed. SERAC~\citep{mitchell2022serac} uses an edit cache, and IKE~\citep{zheng2023ike} uses in-context demonstrations. Larimar~\citep{das2024larimar}, WISE~\citep{wang2024wise}, and NeuralDB~\citep{fei2025neuraldb} use additional weights or memory components.
2) \emph{Parameter-modifying methods.} More directly related to our work, these methods update existing model parameters. Constrained fine-tuning~\citep{zhu2020modifying} limits parameter changes during optimization, while MEND~\citep{mitchell2022mend} learns to transform editing gradients into weight updates. ROME~\citep{meng2022rome} and MEMIT~\citep{meng2023memit} use locate-then-edit to first compute target FFN outputs for new facts, then solve for the corresponding weight updates. UnKE~\citep{deng2025unke} extends parameter-modifying editing to unstructured knowledge. To preserve unrelated knowledge during parameter updates, AlphaEdit~\citep{fang2025alphaedit} projects parameter changes into the null space of a preservation set. For different model architectures, STEM~\citep{sadhukhan2026stem} revises facts through token-indexed embedding substitution, while MoEEdit~\citep{gu2026moeedit} jointly updates MoE experts with constraints on routing shifts. EngramEdit focuses on realizing the potential of pretrained conditional memory for decoupled knowledge updates.

\section{Conclusion and Future Work}
\label{sec:conclusion}

In this work, we propose EngramEdit for decoupled knowledge updates via conditional memory, jointly updating shared $n$-gram embeddings with reuse-based regularization. Experiments show high editing success and generalization, multi-hop use of updated facts, and largely preserved unrelated knowledge and general capabilities. These findings show that EngramEdit enables decoupled knowledge updates in LLMs through conditional memory, extending its role beyond model scaling to an editable knowledge interface. Future work can explore other conditional memory architectures and model scales, and efficient, stable updates to more facts at higher frequencies.

\subsection*{AI use statement}

In this work, we used generative AI tools to assist with method implementation, proof drafting, manuscript writing and revision, and figure creation. AI tools also assisted in discussing and refining the interpretation of experimental results. The authors reviewed the method implementation and proofs and tested the code. We take full responsibility for the final content of this work, including all AI-assisted text, claims, and artifacts.

\subsection*{Ethics statement}

EngramEdit is intended to help LLMs keep factual knowledge up to date, but the ability to revise knowledge can also be misused to introduce false or misleading content. Our evaluation uses existing benchmarks, where counterfactual edit targets serve as experimental test cases and should not be treated as verified real-world facts. Responsible use therefore requires verifying factual updates, restricting editing access to authorized users, and auditing intended and unintended effects.

\subsection*{Reproducibility statement}

All the results in this work are reproducible. Section~\ref{sec:method} describes the method, with the complete editing procedure, theoretical assumptions, and proofs provided in Appendix~\ref{app:method_details_and_proofs}. Section~\ref{sec:experimental_setup} and Appendix~\ref{app:experimental_setup} document the datasets, evaluation protocols, and implementation settings. The code is available in our \href{\CodeRepository}{repository}.

\bibliographystyle{unsrtnat}
\bibliography{ref}

\clearpage
\appendix
\section{Method Details and Proofs}
\label{app:method_details_and_proofs}

This section provides the background formulation, additional method details, and proofs referenced in Sections~\ref{sec:preliminaries} and~\ref{sec:method}.

\subsection{Locate-Then-Edit Formulation}
\label{app:locate_then_edit}

Locate-then-edit first identifies an internal computation associated with the factual association being edited, then determines the output that this computation should produce for the desired prediction, and finally updates the corresponding parameters to reproduce that output. We illustrate this process using a single FFN output projection. Specific editors may use different layer-selection strategies, preservation constraints, or solvers~\citep{meng2022rome,meng2023memit,fang2025alphaedit,gu2026moeedit}.

\paragraph{Causal Localization.}
For the edit $e_i=(s_i,r_i,o_i\rightarrow o_i^\star)$, causal tracing measures how strongly an internal state contributes to recalling the original object $o_i$~\citep{meng2022rome}. The procedure first runs the edit prompt $x_i$ with the subject-token embeddings corrupted. Let $p_i^{\mathrm{corr}}$ denote the resulting probability assigned to $o_i$. It then restores the clean hidden state at token position $q$ and layer $\ell$, retaining the input corruption and recomputing downstream states. Let $p_i^{\mathrm{restore}}(\ell,q)$ denote the probability after this restoration. The indirect effect of the restored state is
\begin{equation}
    \operatorname{IE}_i(\ell,q)
    =p_i^{\mathrm{restore}}(\ell,q)-p_i^{\mathrm{corr}}.
    \label{eq:appendix_indirect_effect}
\end{equation}
A large indirect effect indicates that the state at $(\ell,q)$ contributes strongly to recalling the fact. This analysis identifies, or motivates a fixed choice of, the FFN layer $\ell$ and subject-token position $q_i$ used for editing. At the selected location, let $\mathbf{k}_i^\ell$ be the post-nonlinearity FFN activation and $\mathbf{W}_{\mathrm{out}}^\ell$ the FFN output projection. In practice, the activation $\mathbf{k}_i^\ell$ can be averaged over prompts that place the same subject in different contexts to reduce dependence on one prompt~\citep{meng2022rome,meng2023memit}. The corresponding FFN output is
\begin{equation}
    \mathbf{v}_i^\ell
    =\mathbf{W}_{\mathrm{out}}^\ell\mathbf{k}_i^\ell.
    \label{eq:appendix_localized_ffn_output}
\end{equation}
The remaining two stages determine the output that should replace $\mathbf{v}_i^\ell$ and the parameter update that realizes this replacement.

\paragraph{Target Output Computation.}
Target output computation asks what FFN output at the localized position would make the model predict the desired object $o_i^\star$. Let $\mathcal{C}_i$ contain the edit prompt $x_i$ and any context-augmented prompts used during this computation. With all model parameters fixed, a learnable perturbation $\boldsymbol{\delta}_i$ is temporarily added to the FFN output at the selected position for every prompt in $\mathcal{C}_i$. We denote the model evaluated with this temporary perturbation by $\mathcal{M}_{\boldsymbol{\delta}_i}$. The perturbation is obtained by minimizing
\begin{equation}
\begin{aligned}
    \boldsymbol{\delta}_i^\star
    =\arg\min_{\boldsymbol{\delta}_i}\;&
    \frac{1}{|\mathcal{C}_i|}
    \sum_{x\in\mathcal{C}_i}
    \mathcal{L}_{\mathrm{NLL}}
    \bigl(\mathcal{M}_{\boldsymbol{\delta}_i}(x),o_i^\star\bigr) \\
    &+\lambda_{\mathrm{KL}}
    D_{\mathrm{KL}}
    \bigl(p_{\boldsymbol{\delta}_i}^{\mathrm{pre}}\,\|\,p_0^{\mathrm{pre}}\bigr),
\end{aligned}
\label{eq:appendix_ffn_target_optimization}
\end{equation}
where $p_0^{\mathrm{pre}}$ and $p_{\boldsymbol{\delta}_i}^{\mathrm{pre}}$ are the predictive distributions on a preservation prompt before and during the temporary perturbation, respectively~\citep{meng2022rome}. The NLL term promotes the desired prediction, while the KL term limits changes to the preservation distribution. The optimized perturbation defines the target FFN output
\begin{equation}
    \mathbf{v}_i^\star
    =\mathbf{v}_i^\ell+\boldsymbol{\delta}_i^\star.
    \label{eq:appendix_target_ffn_output}
\end{equation}
The perturbation $\boldsymbol{\delta}_i^\star$ is used only to determine this target. It is removed after optimization and is not written to the model.

\paragraph{Parameter Update.}
The final stage changes the FFN output projection so that it produces the target output from the same activation. For a batch of $m$ edits applied to layer $\ell$, define the activation matrix $\mathbf{K}=[\mathbf{k}_1^\ell,\ldots,\mathbf{k}_m^\ell]$ and the target output matrix $\mathbf{V}^\star=[\mathbf{v}_1^\star,\ldots,\mathbf{v}_m^\star]$. The parameter update $\Delta\mathbf{W}_{\mathrm{out}}^\ell$ should satisfy
\begin{equation}
    (\mathbf{W}_{\mathrm{out}}^\ell+\Delta\mathbf{W}_{\mathrm{out}}^\ell)
    \mathbf{K}
    \approx\mathbf{V}^\star.
    \label{eq:appendix_batch_target_matching}
\end{equation}
Matching only the new targets can change outputs associated with other FFN activations. Let $\mathbf{K}_0$ contain preservation activations sampled from the model. Because their original outputs are $\mathbf{W}_{\mathrm{out}}^\ell\mathbf{K}_0$, preserving these associations amounts to keeping $\Delta\mathbf{W}_{\mathrm{out}}^\ell\mathbf{K}_0$ small. With nonnegative coefficients $\lambda_{\mathrm{preserve}}$ and $\lambda_{\mathrm{ridge}}$, a regularized batch formulation is
\begin{equation}
\begin{aligned}
    \Delta\mathbf{W}_{\mathrm{out}}^{\ell\star}
    =\arg\min_{\Delta\mathbf{W}_{\mathrm{out}}^\ell}\;&
    \left\|
        (\mathbf{W}_{\mathrm{out}}^\ell+\Delta\mathbf{W}_{\mathrm{out}}^\ell)
        \mathbf{K}-\mathbf{V}^\star
    \right\|_F^2 \\
    &+\lambda_{\mathrm{preserve}}
    \left\|\Delta\mathbf{W}_{\mathrm{out}}^\ell\mathbf{K}_0\right\|_F^2
    +\lambda_{\mathrm{ridge}}
    \left\|\Delta\mathbf{W}_{\mathrm{out}}^\ell\right\|_F^2.
\end{aligned}
\label{eq:appendix_ffn_update_objective}
\end{equation}
Define the target residual matrix and the regularized covariance matrix as
\begin{equation}
    \mathbf{R}
    =\mathbf{V}^\star-\mathbf{W}_{\mathrm{out}}^\ell\mathbf{K},
    \qquad
    \mathbf{C}_0
    =\lambda_{\mathrm{preserve}}\mathbf{K}_0\mathbf{K}_0^\top
    +\lambda_{\mathrm{ridge}}\mathbf{I}.
    \label{eq:appendix_ffn_update_terms}
\end{equation}
When $\mathbf{K}\mathbf{K}^\top+\mathbf{C}_0$ is invertible, setting the derivative of Eq.~\ref{eq:appendix_ffn_update_objective} to zero gives
\begin{equation}
    \Delta\mathbf{W}_{\mathrm{out}}^{\ell\star}
    =\mathbf{R}\mathbf{K}^\top
    \left(\mathbf{K}\mathbf{K}^\top+\mathbf{C}_0\right)^{-1},
    \label{eq:appendix_ffn_batch_solution}
\end{equation}
which jointly writes a batch of target FFN outputs while regularizing changes on preservation activations~\citep{meng2023memit}. A positive $\lambda_{\mathrm{ridge}}$ ensures this invertibility.

For a single edit, ROME instead requires exact target matching and minimizes the preservation error under this constraint~\citep{meng2022rome}. With $\mathbf{C}_0=\mathbf{K}_0\mathbf{K}_0^\top$ invertible and $\mathbf{k}_i^\ell\ne\mathbf{0}$, the resulting rank-one update is
\begin{equation}
    \Delta\mathbf{W}_{\mathrm{out}}^{\ell\star}
    =
    \frac{
        (\mathbf{v}_i^\star-\mathbf{W}_{\mathrm{out}}^\ell\mathbf{k}_i^\ell)
        (\mathbf{C}_0^{-1}\mathbf{k}_i^\ell)^\top
    }{
        (\mathbf{C}_0^{-1}\mathbf{k}_i^\ell)^\top\mathbf{k}_i^\ell
    }.
    \label{eq:appendix_ffn_rank_one_solution}
\end{equation}
Later editors modify the preservation constraint, the parameter blocks being updated, or the batch solver, while retaining the separation between target output computation and persistent parameter updating~\citep{fang2025alphaedit,gu2026moeedit}.

\subsection{Conditional Memory Target Computation}
\label{app:target_computation_details}

We provide the prompt for expression generation and the full objective for target computation in Section~\ref{sec:conditional_memory_target_computation}.

\subsubsection{Expression Generation}
\label{app:expression_generation}

To cover different $n$-grams activated at the last subject token, we ask the model to vary the wording immediately before the subject. Figure~\ref{fig:expression_generation_prompt} shows the prompt used for CounterFact and each atomic edit in MQuAKE. Here, \texttt{<original stem>} is the edit prompt $x_i$ with the subject $s_i$ replaced by \texttt{[SUBJ]}; \texttt{<old answer>} and \texttt{<new answer>} are the original and desired objects. For ZsRE, the instruction instead requests a question with the same meaning and answer type.

\begin{figure}[t]
\begin{tcolorbox}[
    enhanced,
    arc=2.5mm,
    boxrule=0.6pt,
    colframe=promptaccent,
    colback=promptaccent!3!white,
    colbacktitle=promptaccent!12!white,
    coltitle=promptaccent!45!black,
    title={Expression Generation Prompt},
    fonttitle=\sffamily\bfseries\fontsize{9}{11}\selectfont,
    fontupper=\ttfamily\fontsize{8}{10}\selectfont,
    titlerule=0pt,
    left=10pt,
    right=10pt,
    top=7pt,
    bottom=7pt,
    toptitle=5pt,
    bottomtitle=5pt,
    before skip=8pt,
    after skip=8pt
]
\raggedright
\setlength{\parskip}{2pt}
Rewrite the factual prompt stem below into one alternative stem with the same meaning.\par
Rules:\par
- The output must contain [SUBJ] exactly once.\par
- Keep it as an incomplete factual sentence stem that naturally expects a completion.\par
- Keep the meaning aligned with the original stem.\par
- Prioritize changing the local phrasing immediately before [SUBJ] when natural.\par
- Good ways to vary it include changing the relation phrase or preposition before [SUBJ], using a possessive form like [SUBJ]'s ..., or rewriting as a question.\par
- Do not rely only on changing words after [SUBJ]; try to make the left-side local frame meaningfully different.\par
- Natural paraphrases are more important than forcing an unnatural token pattern.\par
- Do not use the answer words '<old answer>' or '<new answer>'.\par
- Do not use placeholders like ... or \_\_\_\_.\par
- Do not add numbering, explanation, quotation marks, or multiple options.\par
- Output exactly one rewritten stem and nothing else.\par
\medskip
Original stem: <original stem>
\end{tcolorbox}
\caption{Expression-generation prompt for CounterFact and atomic MQuAKE edits. \texttt{[SUBJ]} marks the subject; angle-bracketed fields are filled from the edit request.}
\label{fig:expression_generation_prompt}
\end{figure}

The instruction is passed as a user message through the model's chat template. We sample candidates in batches and filter duplicates, malformed outputs, and outputs containing answer strings. Later rounds append accepted expressions and ask the model to vary their wording before \texttt{[SUBJ]}. Generation stops once enough expressions are accepted or the round limit is reached. We restore the original subject in each accepted expression before target computation.

\subsubsection{Target Computation Objective}
\label{app:target_computation_objective}

We compute target memory representations by optimizing a temporary perturbation $\boldsymbol{\delta}_i$ using the prediction loss~\citep{meng2023memit}. The perturbation is shared across the expression set $\mathcal{X}_i$ (see Section~\ref{sec:conditional_memory_target_computation}). For a desired object with tokens $o_i^\star=(o_{i,1}^\star,\ldots,o_{i,L_i}^\star)$, the NLL term in Eq.~\ref{eq:target_optimization} is
\begin{equation}
    \mathcal{L}_{\mathrm{NLL}}
    \bigl(\mathcal{M}_{\boldsymbol{\delta}_i}(x),o_i^\star\bigr)
    =-\frac{1}{L_i}\sum_{r=1}^{L_i}
    \log p_{\mathcal{M}_{\boldsymbol{\delta}_i}}
    \bigl(o_{i,r}^\star\mid x,o_{i,<r}^\star\bigr).
    \label{eq:appendix_target_nll}
\end{equation}
Here, $o_{i,<r}^\star$ denotes the preceding tokens of the desired object; the loss averages over its $L_i$ tokens.

The target regularizer combines a preservation loss with a norm penalty~\citep{meng2022rome}:
\begin{equation}
    \mathcal{R}_{\mathrm{target}}(\boldsymbol{\delta}_i)
    =\lambda_{\mathrm{KL}}
    D_{\mathrm{KL}}
    \bigl(p_{\boldsymbol{\delta}_i}^{\mathrm{pre}}\,\|\,p_0^{\mathrm{pre}}\bigr)
    +\lambda_{\mathrm{norm}}
    \frac{\|\boldsymbol{\delta}_i\|_2}{\|\mathbf{h}_i(x_i)\|_2^2},
    \label{eq:appendix_target_regularizer}
\end{equation}
where $p_0^{\mathrm{pre}}$ and $p_{\boldsymbol{\delta}_i}^{\mathrm{pre}}$ are the predictive distributions before and during target computation on the preservation prompt ``$s_i$ is a''. The KL term limits prediction changes when describing the subject. The norm penalty discourages large perturbations, using the original edit prompt's memory representation $\mathbf{h}_i(x_i)$ for normalization. The clamp factor $\rho$ further bounds the perturbation norm through $\|\boldsymbol{\delta}_i\|_2\leq\rho\|\mathbf{h}_i(x_i)\|_2$.

Adding the optimized perturbation $\boldsymbol{\delta}_i^\star$ to each current memory representation $\mathbf{h}_i(x)$ gives its target $\mathbf{h}_i^\star(x)$. The perturbation is then removed; only the $n$-gram embedding updates derived below are written to conditional memory. Appendix~\ref{app:implementation_target_computation} gives the prefix contexts and optimization settings used in our experiments.

\subsection{Joint Memory Update Allocation}
\label{app:memory_update_construction}

To realize the target memory representations, we express how embedding updates change the memory computation. Sum aggregation gives a linear matching relation, while context-aware gating requires accounting for changes in the gate. We derive these two cases and then specify the reuse-based regularization weights used in Section~\ref{sec:joint_memory_updating}.

\subsubsection{Sum Aggregation}
\label{app:sum_aggregation}
We first derive how embedding updates combine to match each target under sum aggregation. LongCat-Flash-Lite uses additive fusion by averaging projected $n$-gram embeddings with the token embedding~\citep{liu2026scalingembeddings}. Here, we use the simplified sum formulation from Section~\ref{sec:joint_memory_updating}; Appendix~\ref{app:implementation_memory_updating} describes how updates are stored and applied in our implementation. For row $p$ of the mapping matrix $\mathbf{A}$, the current memory representation is
\begin{equation}
    \mathbf{h}_p
    =\mathbf{c}_p+
    \sum_{j=1}^{Q}A_{pj}\mathbf{e}(g_j),
    \label{eq:appendix_additive_aggregation}
\end{equation}
where $\mathbf{c}_p$ contains the components that are not updated. Since these components remain fixed, applying the embedding updates $\{\mathbf{u}_j\}_{j=1}^{Q}$ changes only the summed embeddings:
\begin{equation}
    \widetilde{\mathbf{h}}_p(\mathbf{U})
    =\mathbf{h}_p+
    \sum_{j=1}^{Q}A_{pj}\mathbf{u}_j.
    \label{eq:appendix_updated_memory_representation}
\end{equation}
The fixed components cancel when we subtract the current representation from its target. Matching the target memory representation $\mathbf{h}_p^\star$ therefore requires
\begin{equation}
    \sum_{j=1}^{Q}A_{pj}\mathbf{u}_j
    \approx
    \mathbf{h}_p^\star-\mathbf{h}_p
    =\mathbf{b}_p.
    \label{eq:appendix_row_matching}
\end{equation}
Stacking these relations over the edit batch gives $\mathbf{A}\mathbf{U}\approx\mathbf{B}$, so one embedding update is jointly constrained by all expressions that use it.

\subsubsection{Context-Aware Gated Aggregation}
\label{app:gated_aggregation}
To extend target matching to context-aware gating, we account for how embedding updates affect both the projected values and their gates. Engram~\citep{cheng2026engram}, Qwen3.8-Flash-Next~\citep{qwen2026design}, and DeepSeek-V4.1-Flash~\citep{deepseekai2026deepseekv41flash} use contextual gating to control the contribution of $n$-gram embeddings. We illustrate the matching relation using Engram's scalar gate.

For expression $x_p$, let $\mathbf{z}_p\in\mathbb{R}^{d_z}$ concatenate the activated embeddings and $\mathbf{q}_p\in\mathbb{R}^{d}$ denote the contextual query supplied by the backbone. Here, $d_z$ is the dimension of the concatenated embedding vector. We first consider updates that leave the upstream query computation unchanged, so $\mathbf{q}_p$ remains fixed. The memory function $\mathbf{F}_p$ and gate $\alpha_p$ are
\begin{equation}
    \mathbf{h}_p
    =\mathbf{F}_p(\mathbf{z}_p;\mathbf{q}_p),
    \qquad
    \alpha_p
    =\sigma\!\left(
        \frac{
        \operatorname{RMSNorm}(\mathbf{q}_p)^{\top}
        \operatorname{RMSNorm}(\mathbf{W}_K\mathbf{z}_p)
        }{\sqrt{d}}
    \right),
    \label{eq:appendix_gated_aggregation}
\end{equation}
where $\sigma$ is the sigmoid function and $\mathbf{W}_K,\mathbf{W}_V\in\mathbb{R}^{d\times d_z}$ are fixed key and value projection matrices. The memory function $\mathbf{F}_p$ includes the gated value $\alpha_p\mathbf{W}_V\mathbf{z}_p$ and any subsequent architecture-specific processing, whose parameters also remain fixed.

To apply the same update wherever an embedding is shared, we use the batch-level embedding update matrix $\mathbf{U}$. Its vectorization $\operatorname{vec}(\mathbf{U})\in\mathbb{R}^{Qd}$ stacks its columns. The selection operator $\mathbf{S}_p\in\mathbb{R}^{d_z\times Qd}$ places each update wherever its embedding occurs in $\mathbf{z}_p$, giving
\begin{equation}
    \widetilde{\mathbf{h}}_p(\mathbf{U})
    =\mathbf{F}_p
    \left(
        \mathbf{z}_p+\mathbf{S}_p\operatorname{vec}(\mathbf{U});
        \mathbf{q}_p
    \right).
    \label{eq:appendix_updated_gated_aggregation}
\end{equation}
Local processing such as a causal convolution can make the memory representation depend on earlier token positions. In that case, the mapping matrix $\mathbf{A}$ and selection operator $\mathbf{S}_p$ include the activated $n$-grams throughout its receptive field. To match targets after the gated computation, we replace the additive matching term in Eq.~\ref{eq:joint_memory_update} with
\begin{equation}
    \sum_{p=1}^{P}
    \left\|
        \widetilde{\mathbf{h}}_p(\mathbf{U})-\mathbf{h}_p^\star
    \right\|_2^2,
    \label{eq:appendix_nonlinear_gated_objective}
\end{equation}
while retaining the same reuse-based regularization. The resulting objective can be optimized with gradients.

To use a linear solver, we can instead approximate the gated computation locally at the current embeddings. Let $\mathbf{J}_p\in\mathbb{R}^{d\times d_z}$ be the Jacobian of $\mathbf{F}_p$ with respect to $\mathbf{z}_p$, with the query $\mathbf{q}_p$ fixed. A first-order approximation gives
\begin{equation}
    \widetilde{\mathbf{h}}_p(\mathbf{U})-\mathbf{h}_p
    \approx
    \mathbf{J}_p\mathbf{S}_p\operatorname{vec}(\mathbf{U}).
    \label{eq:appendix_gated_linearization}
\end{equation}
Stacking these relations gives a regularized linear system that approximates the gated objective, using the same matching targets $\mathbf{b}_p=\mathbf{h}_p^\star-\mathbf{h}_p$ and regularization coefficients $\lambda_j$ as the additive formulation. If embedding updates also affect the upstream computation, the contextual query $\mathbf{q}_p$ is recomputed during optimization, and the Jacobian includes its dependence on the updates. Thus, changing the aggregation modifies the matching relation and solver while retaining target computation, memory mapping, and reuse-based regularization.

\subsubsection{Regularization Weight Construction}
\label{app:protection_weight}
For either aggregation, we penalize updates to frequently reused embeddings more strongly to limit unintended changes on unrelated inputs. Shorter $n$-grams can be shared by more inputs, while corpus frequency measures how often each $n$-gram occurs. We therefore use length and frequency as indicators of reuse. For the $n$-gram $g_j$, let $n_j$ denote its length and $\pi_j\in[0,1]$ its within-length corpus-frequency percentile. The regularization weight $w_j$ combines these two indicators:
\begin{equation}
\begin{aligned}
    w_j^{\mathrm{len}} &= \eta_{n_j}, \\
    w_j^{\mathrm{freq}} &=
    \min\!\left(1+\gamma\pi_j^{\beta},\,w_{\mathrm{freq}}^{\max}\right), \\
    w_j &=
    \min\!\left(
        w_j^{\mathrm{len}}+w_j^{\mathrm{freq}}-1,
        w^{\max}
    \right),
\end{aligned}
\label{eq:reuse_weight}
\end{equation}
where $\eta_n>0$ decreases with $n$ to assign stronger penalties to shorter $n$-grams. The coefficient $\gamma\geq0$ and exponent $\beta>0$ control the strength and shape of frequency weighting. Subtracting 1 makes frequency weighting an increment over the length weight. The caps $w_{\mathrm{freq}}^{\max}\geq1$ and $w^{\max}>0$ prevent either the frequency weight or the combined weight from becoming excessively large.

To control the overall penalty strength, we set the regularization coefficient $\lambda_j=\lambda_{\mathrm{reuse}}w_j+\lambda_{\mathrm{ridge}}$. The coefficient $\lambda_{\mathrm{reuse}}$ scales the reuse-based weights; the optional $\lambda_{\mathrm{ridge}}$ adds a uniform penalty independently of length and frequency. Setting $\lambda_{\mathrm{ridge}}=0$ removes this uniform term. Both coefficients are nonnegative and chosen so that every $\lambda_j>0$. Under sum aggregation, Appendix~\ref{app:reuse_protection} relates length and frequency to an upper bound on memory representation changes for unrelated inputs, providing a theoretical basis for these weights.

\subsection{Complete Editing Procedure}
\label{app:complete_editing_procedure}

Algorithm~\ref{alg:engramedit} summarizes EngramEdit for one edit batch under sum aggregation. Target computation precedes memory mapping, and the selected $n$-gram embeddings are updated only after solving the batch-level joint problem. For context-aware gated aggregation, the matching objective and solver are replaced as described in Appendix~\ref{app:gated_aggregation}.

\begin{algorithm}[t]
\caption{EngramEdit for One Edit Batch}
\label{alg:engramedit}
\begin{algorithmic}[1]
\Require Current model $\mathcal{M}$ and edit batch $\mathcal{E}=\{e_i\}_{i=1}^{m}$
\Require Number of generated expressions $K$ and configured $n$-gram lengths $\mathcal{N}$
\Require Regularization coefficients $\lambda_{\mathrm{reuse}},\lambda_{\mathrm{ridge}}\geq0$ and clamp factor $\rho>0$
\Require $\lambda_j=\lambda_{\mathrm{reuse}}w_j+\lambda_{\mathrm{ridge}}>0$ for each selected embedding
\Ensure Edited model $\mathcal{M}'$ with an updated conditional memory
\ForAll{$e_i=(s_i,r_i,o_i\rightarrow o_i^\star)\in\mathcal{E}$}
    \State $x_i\gets$ original edit prompt for $(s_i,r_i)$ \Comment{Expression Generation}
    \State $\mathcal{P}_i\gets\operatorname{GenerateExpressions}(\mathcal{M},e_i,K)$
    \State $\mathcal{X}_i\gets\{x_i\}\cup\mathcal{P}_i$
    \ForAll{$x\in\mathcal{X}_i$} \Comment{Joint Target Computation}
        \State $q_i(x)\gets$ last subject-token position in $x$
        \State $\mathbf{h}_i(x)\gets\mathbf{h}(x,q_i(x))$
    \EndFor
    \State Compute the shared perturbation $\boldsymbol{\delta}_i^\star$ using Eq.~\ref{eq:target_optimization} with clamp factor $\rho$
    \State $\mathbf{h}_i^\star(x)\gets\mathbf{h}_i(x)+\boldsymbol{\delta}_i^\star$ for all $x\in\mathcal{X}_i$
\EndFor
\State $\mathcal{R}\gets\{(i,x):e_i\in\mathcal{E},\;x\in\mathcal{X}_i\}$
\State $\mathcal{G}_i(x)\gets\mathcal{G}(x,q_i(x))$ for all $(i,x)\in\mathcal{R}$ \Comment{Memory Mapping}
\State Enumerate $\mathcal{R}$ as $\{(i_p,x_p)\}_{p=1}^{P}$, where $P=|\mathcal{R}|$
\State $\{g_j\}_{j=1}^{Q}\gets
\operatorname{Unique}\bigl(\bigcup_{(i,x)\in\mathcal{R}}\mathcal{G}_i(x)\bigr)$
\State $A_{pj}\gets\mathbb{I}[g_j\in\mathcal{G}_{i_p}(x_p)]$ for all $p,j$
\Comment{Mapping Matrix}
\State $\mathbf{b}_p\gets\boldsymbol{\delta}_{i_p}^\star$ for all $p$;
$\mathbf{B}\gets[\mathbf{b}_1,\ldots,\mathbf{b}_P]^\top$ \Comment{Target Matching}
\State Compute $w_j$ using Eq.~\ref{eq:reuse_weight} and set
$\lambda_j\gets\lambda_{\mathrm{reuse}}w_j+\lambda_{\mathrm{ridge}}$
\State $\boldsymbol{\Lambda}\gets\operatorname{diag}(\lambda_1,\ldots,\lambda_Q)$ \Comment{Reuse-Based Regularization}
\State $\mathbf{U}^\star\gets
\operatorname{Solve}(\mathbf{A}^{\top}\mathbf{A}+\boldsymbol{\Lambda},
\mathbf{A}^{\top}\mathbf{B})$
\Comment{Joint Update Solution}
\ForAll{selected $n$-grams $g_j$} \Comment{Memory Updating}
    \State $\mathbf{e}(g_j)\gets\mathbf{e}(g_j)+\mathbf{u}_j^\star$
\EndFor
\State \Return $\mathcal{M}'$
\end{algorithmic}
\end{algorithm}

\paragraph{Sequential Editing.}
Sequential editing repeats this procedure on the current conditional memory. At step $t$, both the target memory representations and the matrices $\mathbf{B}_t$ and $\boldsymbol{\Lambda}_t$ are computed from the model after steps $1,\ldots,t-1$. The solved update is then added to the current $n$-gram embedding. This procedure accounts for previous changes whenever a later edit activates an $n$-gram that has already been updated.

\paragraph{Computational Cost.}
Let $s$ be the maximum number of configured $n$-grams activated by one expression. For an edit batch with $P$ expressions, $Q$ unique $n$-grams, and embedding dimension $d$, forming the sparse normal equations costs $O(Ps^2+Psd)$. Solving the resulting dense $Q\times Q$ system costs $O(Q^3+Q^2d)$ and uses $O(Q^2+Qd)$ memory, excluding model storage. If no selected $n$-gram embedding is shared between groups of expressions, the system can be split into smaller systems. Solving them separately preserves the joint solution and can reduce solver time and memory (see Appendix~\ref{app:component_decomposition} for the decomposition proof).

\subsection{Proofs for Joint Memory Update Allocation}
\label{app:joint_update_proofs}

We analyze the joint update in Section~\ref{sec:joint_memory_updating} under sum aggregation (see Appendix~\ref{app:sum_aggregation}). The results explain the need for approximate matching, support reuse-based regularization, and characterize the solution's update cost, sensitivity, and computation.

We use the mapping matrix $\mathbf{A}\in\mathbb{R}^{P\times Q}$, matching target matrix $\mathbf{B}\in\mathbb{R}^{P\times d}$, and embedding update matrix $\mathbf{U}\in\mathbb{R}^{Q\times d}$. The $j$th row of $\mathbf{U}$ is $\mathbf{u}_j^\top$. For the regularized objective, $\boldsymbol{\Lambda}=\operatorname{diag}(\lambda_1,\ldots,\lambda_Q)$ with every $\lambda_j>0$. We write this objective as
\begin{equation}
    F(\mathbf{U})
    =\|\mathbf{A}\mathbf{U}-\mathbf{B}\|_F^2
    +\|\boldsymbol{\Lambda}^{1/2}\mathbf{U}\|_F^2,
    \label{eq:appendix_joint_objective}
\end{equation}
and let $\mathbf{U}^\star$ denote a minimizer. The notation $\|\cdot\|_F$ denotes the Frobenius norm; $\|\cdot\|_2$ denotes the Euclidean norm for vectors and the spectral norm for matrices.

\subsubsection{Feasibility of Exact Matching}
\label{app:matching_structure}
Shared embeddings can make matching targets incompatible. For example, expressions that activate the same selected $n$-grams receive the same memory representation change. If their matching targets differ, no embedding update can satisfy both exactly. The following condition identifies when exact matching is possible.

\begin{propositionbox}
\textbf{Proposition 1.}
Exact matching $\mathbf{A}\mathbf{U}=\mathbf{B}$ is possible if and only if every column of the matching target matrix $\mathbf{B}$ lies in the column space of the mapping matrix $\mathbf{A}$. Equivalently,
\begin{equation}
    \mathbf{z}^{\top}\mathbf{B}=\mathbf{0}^{\top}
    \quad\text{for every }\mathbf{z}\in\operatorname{null}(\mathbf{A}^{\top}).
    \label{eq:appendix_matching_feasibility}
\end{equation}
\end{propositionbox}

\paragraph{Proof.}
Exact matching separates into one equation per embedding dimension, $\mathbf{A}\mathbf{U}_{:,r}=\mathbf{B}_{:,r}$ for $r=1,\ldots,d$. Each equation has a solution precisely when its target column belongs to $\operatorname{col}(\mathbf{A})$. Stacking these column solutions gives an exact update matrix. The identity $\operatorname{col}(\mathbf{A})=\operatorname{null}(\mathbf{A}^{\top})^{\perp}$ gives the equivalent condition in Eq.~\ref{eq:appendix_matching_feasibility}.

When exact matching is infeasible, the squared matching error in Eq.~\ref{eq:joint_memory_update} allows a joint compromise across conflicting targets. The proposition concerns feasibility; positive regularization can also favor an inexact match when an exact match exists.

\subsubsection{Reuse-Based Regularization}
\label{app:reuse_protection}
\label{app:reuse_bound}
An embedding update can also change memory representations on unrelated inputs that activate the same $n$-gram. We relate these changes to activation probabilities to explain why frequently reused embeddings receive stronger update penalties.

Let $\mathcal{D}$ be a fixed distribution over unrelated input--position pairs $(x,q)$. For each selected $n$-gram $g_j$, define its activation indicator $a_j(x,q)=\mathbb{I}[g_j\in\mathcal{G}(x,q)]$ and activation probability $p_j=\mathbb{E}_{\mathcal{D}}[a_j(x,q)]$. Let $s\leq|\mathcal{N}|$ be the maximum number of selected $n$-grams activated at one position under the lookup in Section~\ref{sec:preliminaries}. For fixed embedding updates,
\begin{equation}
    \widetilde{\mathbf{h}}(x,q;\mathbf{U})-\mathbf{h}(x,q)
    =\sum_{j=1}^{Q}a_j(x,q)\mathbf{u}_j.
    \label{eq:appendix_unrelated_memory_change}
\end{equation}

\begin{propositionbox}
\textbf{Proposition 2.}
For any fixed embedding updates $\{\mathbf{u}_j\}_{j=1}^{Q}$, the expected squared memory representation change satisfies
\begin{equation}
    \mathbb{E}_{\mathcal{D}}\!\left[
        \|\widetilde{\mathbf{h}}(x,q;\mathbf{U})-\mathbf{h}(x,q)\|_2^2
    \right]
    \leq s\sum_{j=1}^{Q}p_j\|\mathbf{u}_j\|_2^2.
    \label{eq:appendix_reuse_change_bound}
\end{equation}
No independence assumption between $n$-gram activations is required.
\end{propositionbox}

\paragraph{Proof.}
Fix an input--position pair $(x,q)$ and let $\mathcal{J}(x,q)=\{j:a_j(x,q)=1\}$ index its activated selected $n$-grams. The triangle and Cauchy--Schwarz inequalities give
\begin{equation}
\begin{aligned}
    \left\|\sum_{j=1}^{Q}a_j(x,q)\mathbf{u}_j\right\|_2^2
    &\leq\left(\sum_{j\in\mathcal{J}(x,q)}\|\mathbf{u}_j\|_2\right)^2\\
    &\leq|\mathcal{J}(x,q)|\sum_{j\in\mathcal{J}(x,q)}\|\mathbf{u}_j\|_2^2
    \leq s\sum_{j=1}^{Q}a_j(x,q)\|\mathbf{u}_j\|_2^2.
\end{aligned}
\label{eq:appendix_pointwise_reuse_bound}
\end{equation}
The bound also holds when no selected $n$-gram is activated, since both sides are zero. Taking expectations and using $\mathbb{E}_{\mathcal{D}}[a_j(x,q)]=p_j$ proves Eq.~\ref{eq:appendix_reuse_change_bound} by linearity of expectation.

The bound assigns a larger cost to the same embedding update when its $n$-gram is activated more often on unrelated inputs. This provides a theoretical basis for penalizing large updates to frequently reused embeddings.

\paragraph{Bounds for the Solved Update.}
The activation bound applies to arbitrary updates. For the updates returned by the joint objective, regularization also limits the total weighted update cost, yielding the following bounds.

\begin{propositionbox}
\textbf{Corollary 1.}
The minimizer $\mathbf{U}^\star$ of Eq.~\ref{eq:appendix_joint_objective} satisfies
\begin{equation}
    \|\mathbf{u}_j^\star\|_2
    \leq\frac{\|\mathbf{B}\|_F}{\sqrt{\lambda_j}}
    \quad\text{for every }j,
    \label{eq:appendix_protection_bound}
\end{equation}
and its expected squared memory representation change on unrelated inputs satisfies
\begin{equation}
    \mathbb{E}_{\mathcal{D}}\!\left[
        \|\widetilde{\mathbf{h}}(x,q;\mathbf{U}^\star)-\mathbf{h}(x,q)\|_2^2
    \right]
    \leq s\left(\max_j\frac{p_j}{\lambda_j}\right)\|\mathbf{B}\|_F^2.
    \label{eq:appendix_regularized_reuse_bound}
\end{equation}
\end{propositionbox}

\paragraph{Proof.}
The zero update is feasible, so optimality and nonnegativity of the matching error give
\begin{equation}
    \sum_{j=1}^{Q}\lambda_j\|\mathbf{u}_j^\star\|_2^2
    \leq F(\mathbf{U}^\star)
    \leq F(\mathbf{0})=\|\mathbf{B}\|_F^2.
    \label{eq:appendix_weighted_update_bound}
\end{equation}
Each summand is nonnegative. Bounding the $j$th term by $\|\mathbf{B}\|_F^2$, dividing by $\lambda_j>0$, and taking square roots gives Eq.~\ref{eq:appendix_protection_bound}. For the second bound, apply Proposition 2 and then use
\begin{equation}
    \sum_{j=1}^{Q}p_j\|\mathbf{u}_j^\star\|_2^2
    \leq\left(\max_j\frac{p_j}{\lambda_j}\right)
        \sum_{j=1}^{Q}\lambda_j\|\mathbf{u}_j^\star\|_2^2
    \leq\left(\max_j\frac{p_j}{\lambda_j}\right)\|\mathbf{B}\|_F^2.
\end{equation}

For fixed matching targets, increasing $\lambda_j$ tightens the bound on its embedding update. With the unrelated-input distribution also fixed, the bound on memory representation changes decreases when the largest ratio $p_j/\lambda_j$ decreases. Thus, the bounds connect stronger penalties on frequently activated embeddings to control of memory representation changes on unrelated inputs.

\paragraph{Length and Frequency.}
The activation probabilities on unrelated inputs are not directly available, so EngramEdit uses $n$-gram length and corpus frequency as indicators. If the selected $n$-gram $g_k$ is a shorter suffix of the selected $n$-gram $g_j$, every activation of $g_j$ also activates $g_k$. Hence,
\begin{equation}
    a_j(x,q)\leq a_k(x,q)
    \quad\Longrightarrow\quad p_j\leq p_k.
    \label{eq:appendix_suffix_reuse}
\end{equation}
This ordering holds for suffix-related $n$-grams, not every pair of different lengths.

For a corpus of $N_{\mathrm{corpus}}$ input--position pairs, let $c_j$ count the positions activating $g_j$. Its activation probability under the uniform empirical distribution is $\widehat{p}_j=c_j/N_{\mathrm{corpus}}$. Counts and empirical activation probabilities therefore have the same ordering. The within-length frequency ranks in Appendix~\ref{app:protection_weight} retain this ordering within each length, but do not estimate $p_j$ on unrelated inputs. These relations motivate the length and frequency factors without establishing optimality of the particular weight formula. The bounds concern memory representations, not downstream predictions.

\subsubsection{Closed-Form Solution and Uniqueness}
\label{app:joint_update_solution_proof}
Overlapping activations can make columns of the mapping matrix linearly dependent. Positive regularization ensures a unique joint update without requiring this matrix to have full column rank.

\begin{propositionbox}
\textbf{Proposition 3.}
If every regularization coefficient satisfies $\lambda_j>0$, the objective in Eq.~\ref{eq:joint_memory_update} has the unique solution in Eq.~\ref{eq:joint_memory_update_solution}.
\end{propositionbox}

\paragraph{Proof.}
Differentiating Eq.~\ref{eq:appendix_joint_objective} gives $\nabla_{\mathbf{U}}F=2\mathbf{A}^{\top}(\mathbf{A}\mathbf{U}-\mathbf{B})+2\boldsymbol{\Lambda}\mathbf{U}$. Setting the gradient to zero yields
\begin{equation}
    (\mathbf{A}^{\top}\mathbf{A}+\boldsymbol{\Lambda})\mathbf{U}
    =\mathbf{A}^{\top}\mathbf{B}.
    \label{eq:appendix_normal_equation}
\end{equation}
For any nonzero vector $\mathbf{z}\in\mathbb{R}^{Q}$,
\begin{equation}
    \mathbf{z}^{\top}(\mathbf{A}^{\top}\mathbf{A}+\boldsymbol{\Lambda})\mathbf{z}
    =\|\mathbf{A}\mathbf{z}\|_2^2+\sum_{j=1}^{Q}\lambda_j z_j^2>0.
\end{equation}
The coefficient matrix is therefore positive definite, making the objective strictly convex. Its unique stationary point is the global minimizer,
\begin{equation}
    \mathbf{U}^\star=(\mathbf{A}^{\top}\mathbf{A}+\boldsymbol{\Lambda})^{-1}\mathbf{A}^{\top}\mathbf{B}.
\end{equation}

\subsubsection{Minimum-Cost Update Allocation}
\label{app:minimum_cost_allocation}
Different embedding updates can produce the same memory representation changes on the edit expressions. Extra update components that leave these changes unchanged do not improve target matching, but can affect unrelated inputs that activate different subsets of embeddings. We show that the joint solution avoids such extra components by minimizing the weighted update cost for its achieved memory representation changes.

\begin{propositionbox}
\textbf{Proposition 4.}
Let $\mathbf{U}^\star$ minimize Eq.~\ref{eq:appendix_joint_objective} with every $\lambda_j>0$. Any alternative embedding update matrix $\mathbf{V}\in\mathbb{R}^{Q\times d}$ satisfying $\mathbf{A}\mathbf{V}=\mathbf{A}\mathbf{U}^\star$ obeys
\begin{equation}
    \|\boldsymbol{\Lambda}^{1/2}\mathbf{V}\|_F^2
    =\|\boldsymbol{\Lambda}^{1/2}\mathbf{U}^\star\|_F^2
    +\|\boldsymbol{\Lambda}^{1/2}(\mathbf{V}-\mathbf{U}^\star)\|_F^2.
    \label{eq:appendix_minimum_cost_allocation}
\end{equation}
Thus, $\mathbf{U}^\star$ uniquely minimizes the weighted update cost among all updates producing the same achieved memory representation changes $\mathbf{A}\mathbf{U}^\star$.
\end{propositionbox}

\paragraph{Proof.}
Let $\mathbf{D}=\mathbf{V}-\mathbf{U}^\star$. The shared matching result gives $\mathbf{A}\mathbf{D}=\mathbf{0}$, and the normal equation gives $\boldsymbol{\Lambda}\mathbf{U}^\star=\mathbf{A}^{\top}(\mathbf{B}-\mathbf{A}\mathbf{U}^\star)$. Together, these identities imply
\begin{equation}
    \operatorname{tr}(\mathbf{D}^{\top}\boldsymbol{\Lambda}\mathbf{U}^\star)
    =\operatorname{tr}\!\left((\mathbf{A}\mathbf{D})^{\top}(\mathbf{B}-\mathbf{A}\mathbf{U}^\star)\right)=0,
    \label{eq:appendix_allocation_orthogonality}
\end{equation}
where $\operatorname{tr}$ denotes the matrix trace. Expanding the weighted cost of $\mathbf{V}=\mathbf{U}^\star+\mathbf{D}$ then gives
\begin{equation}
\begin{aligned}
    \|\boldsymbol{\Lambda}^{1/2}\mathbf{V}\|_F^2
    &=\|\boldsymbol{\Lambda}^{1/2}\mathbf{U}^\star\|_F^2
        +2\operatorname{tr}(\mathbf{D}^{\top}\boldsymbol{\Lambda}\mathbf{U}^\star)
        +\|\boldsymbol{\Lambda}^{1/2}\mathbf{D}\|_F^2\\
    &=\|\boldsymbol{\Lambda}^{1/2}\mathbf{U}^\star\|_F^2
        +\|\boldsymbol{\Lambda}^{1/2}\mathbf{D}\|_F^2.
\end{aligned}
\end{equation}
Since every $\lambda_j>0$, the additional cost is strictly positive unless $\mathbf{D}=\mathbf{0}$, proving minimality and uniqueness under the matching constraint.

The comparison fixes the achieved changes $\mathbf{A}\mathbf{U}^\star$, so it remains valid when exact matching to $\mathbf{B}$ is infeasible. Any additional update with columns in $\operatorname{null}(\mathbf{A})$ increases the weighted cost without improving the matched representations. This is a guarantee on weighted update cost, not on downstream prediction changes.

\subsubsection{Sensitivity to Target Changes}
\label{app:solution_sensitivity}
Target memory representations are computed through optimization and may vary slightly. We therefore examine how these variations affect the solved embedding updates when the mapping and regularization remain fixed. The following bound shows how positive regularization controls this sensitivity.

\begin{propositionbox}
\textbf{Proposition 5.}
For fixed mapping matrix $\mathbf{A}$ and positive definite regularization matrix $\boldsymbol{\Lambda}$, let $\Delta\mathbf{B}$ be a change in the matching target matrix. The corresponding solution change satisfies
\begin{equation}
\begin{aligned}
    \Delta\mathbf{U}^\star
    &=(\mathbf{A}^{\top}\mathbf{A}+\boldsymbol{\Lambda})^{-1}\mathbf{A}^{\top}\Delta\mathbf{B},\\
    \|\Delta\mathbf{U}^\star\|_F
    &\leq\frac{\|\mathbf{A}\|_2}{\lambda_{\min}(\boldsymbol{\Lambda})}\|\Delta\mathbf{B}\|_F.
\end{aligned}
\label{eq:appendix_solution_sensitivity}
\end{equation}
\end{propositionbox}

\paragraph{Proof.}
Write $\mathbf{H}=\mathbf{A}^{\top}\mathbf{A}+\boldsymbol{\Lambda}$ and let $\mathbf{U}^\star(\mathbf{B})$ denote the solution for matching targets $\mathbf{B}$. Subtracting the normal equations for $\mathbf{B}+\Delta\mathbf{B}$ and $\mathbf{B}$ gives
\begin{equation}
    \mathbf{H}\Delta\mathbf{U}^\star=\mathbf{A}^{\top}\Delta\mathbf{B},
    \qquad
    \Delta\mathbf{U}^\star=\mathbf{U}^\star(\mathbf{B}+\Delta\mathbf{B})-\mathbf{U}^\star(\mathbf{B}).
\end{equation}
Multiplying by $\mathbf{H}^{-1}$ proves the exact relation. For every unit vector $\mathbf{z}$, $\mathbf{z}^{\top}\mathbf{H}\mathbf{z}\geq\mathbf{z}^{\top}\boldsymbol{\Lambda}\mathbf{z}\geq\lambda_{\min}(\boldsymbol{\Lambda})$. Thus, $\|\mathbf{H}^{-1}\|_2\leq1/\lambda_{\min}(\boldsymbol{\Lambda})$, and
\begin{equation}
    \|\Delta\mathbf{U}^\star\|_F
    \leq\|\mathbf{H}^{-1}\|_2\|\mathbf{A}^{\top}\|_2\|\Delta\mathbf{B}\|_F
    \leq\frac{\|\mathbf{A}\|_2}{\lambda_{\min}(\boldsymbol{\Lambda})}\|\Delta\mathbf{B}\|_F.
\end{equation}
The final step uses $\|\mathbf{A}^{\top}\|_2=\|\mathbf{A}\|_2$.

For a fixed mapping, the smallest regularization coefficient controls the sensitivity bound even when the mapping matrix is rank deficient. The bound quantifies possible amplification of target changes; it does not require this amplification to be below one.

\subsubsection{Component Decomposition}
\label{app:component_decomposition}
Joint solving is needed only among expressions connected through shared embeddings. When groups share no selected $n$-gram embeddings, solving them separately can reduce computation without changing the joint solution.

\begin{propositionbox}
\textbf{Proposition 6.}
The joint update decomposes exactly over disconnected components of the bipartite graph between expression rows and selected $n$-gram columns, with an edge wherever $A_{pj}\neq0$.
\end{propositionbox}

\paragraph{Proof.}
Ordering rows and columns by connected component puts the mapping matrix into block-diagonal form,
\begin{equation}
    \mathbf{A}=\operatorname{diag}(\mathbf{A}_1,\ldots,\mathbf{A}_C).
    \label{eq:appendix_component_decomposition}
\end{equation}
For component $c$, let $\mathbf{B}_c$ contain its expression targets, and let $\mathbf{U}_c$ and $\boldsymbol{\Lambda}_c$ contain the updates and regularization coefficients for its embeddings. Since $\boldsymbol{\Lambda}$ is diagonal, the objective separates as
\begin{equation}
    F(\mathbf{U})=\sum_{c=1}^{C}
    \left(\|\mathbf{A}_c\mathbf{U}_c-\mathbf{B}_c\|_F^2
    +\|\boldsymbol{\Lambda}_c^{1/2}\mathbf{U}_c\|_F^2\right).
\end{equation}
Each summand depends only on its own embedding updates. Applying Proposition 3 to each component gives
\begin{equation}
    \mathbf{U}_c^\star=(\mathbf{A}_c^{\top}\mathbf{A}_c+\boldsymbol{\Lambda}_c)^{-1}\mathbf{A}_c^{\top}\mathbf{B}_c.
\end{equation}
Stacking these solutions and restoring the original embedding order recovers the joint solution. An isolated expression contributes only a constant matching loss; an isolated embedding receives the zero update. Neither affects other components.

\section{Experimental Setup}
\label{app:experimental_setup}

This section describes the datasets, evaluation, baselines, and implementation details for the experiments in Section~\ref{sec:experiments}.

\subsection{Datasets}
\label{app:datasets}

The editing benchmarks test factual recall and the use of revised facts in reasoning. A separate set of tasks measures general capabilities after editing.

\paragraph{CounterFact.}
CounterFact~\citep{meng2022rome} tests counterfactual changes to factual associations. We use the MultiCounterFact release provided with MEMIT for batched editing~\citep{meng2023memit}. Each case contains an edit request $e_i$ and its prompt (see Section~\ref{sec:preliminaries}), held-out paraphrases, neighborhood prompts about other subjects whose facts should remain unchanged, and generation prompts. Held-out paraphrases are used only for evaluation; generated expressions are used during editing. Table~\ref{tab:editing_examples} illustrates an edit request and its evaluation prompts.

\paragraph{ZsRE.}
ZsRE~\citep{levy2017zsre} represents factual associations as questions and answers. We use the editing evaluation split distributed with MEMIT~\citep{meng2023memit}, taking the first reference answer as the editing target. Each case pairs the edit question with a held-out paraphrase and an unrelated question--answer pair from Natural Questions. The edit therefore requests the supplied factual answer, not a counterfactual replacement. An example question, its target answer, and the evaluation questions are shown in Table~\ref{tab:editing_examples}.

\paragraph{MQuAKE.}
MQuAKE~\citep{zhong2023mquake} tests whether revised facts can be combined to answer multi-hop questions. We use 3,000 counterfactual cases covering two- to four-hop reasoning. Each case contains one or more fact-level edit requests, variants of a multi-hop question, and the answer implied by the revised facts, with accepted aliases. Table~\ref{tab:editing_examples} shows how multiple fact edits change the answer to one multi-hop question.

\paragraph{General-Capability Tasks.}
Following AlphaEdit~\citep{fang2025alphaedit}, we evaluate SST (SST-2)~\citep{socher2013sst} for sentiment classification, MRPC~\citep{dolan2005mrpc} for paraphrase recognition, and CoLA~\citep{warstadt2019cola} for linguistic acceptability. RTE~\citep{bentivogli2009rte5} and NLI~\citep{wang2018glue} test textual entailment, and MMLU~\citep{hendrycks2021mmlu} tests knowledge across multiple subjects. We use 100 examples per task; none are included in the edit requests.

\begin{table}[t]
    \caption{Editing examples from CounterFact, ZsRE, and MQuAKE, with simplified formatting. The ellipsis omits an unrelated prefix from the CounterFact paraphrase.}
    \label{tab:editing_examples}
    \centering
    \small
    \begin{tabular}{@{}p{0.23\linewidth}p{0.72\linewidth}@{}}
        \toprule
        \textbf{Field} & \textbf{Example} \\
        \midrule
        \multicolumn{2}{@{}l}{\textbf{CounterFact}} \\
        Subject and relation & Danielle Darrieux; mother tongue \\
        Requested change & French $\rightarrow$ English \\
        Edit prompt & The mother tongue of Danielle Darrieux is \\
        Held-out paraphrase & \ldots\ Danielle Darrieux spoke the language \\
        Unrelated prompt & The native language of Montesquieu is \\
        Unchanged answer & French \\
        \midrule
        \multicolumn{2}{@{}l}{\textbf{ZsRE}} \\
        Subject & Watts Humphrey \\
        Edit question & What university did Watts Humphrey attend? \\
        Desired answer & Illinois Institute of Technology \\
        Held-out paraphrase & What university did Watts Humphrey take part in? \\
        Unrelated question & Who played Desmond Doss's father in Hacksaw Ridge? \\
        Unchanged answer & Hugo Weaving \\
        \midrule
        \multicolumn{2}{@{}l}{\textbf{MQuAKE}} \\
        Edit request 1 & Fernando Santos; country of citizenship: Portugal $\rightarrow$ United Kingdom \\
        Edit request 2 & United Kingdom; head of state: Elizabeth II $\rightarrow$ Emmerson Mnangagwa \\
        Multi-hop question & Who is the head of state of the country where Fernando Santos hold a citizenship? \\
        Original answer & Marcelo Rebelo de Sousa \\
        Desired answer & Emmerson Mnangagwa \\
        \bottomrule
    \end{tabular}
\end{table}

\subsection{Evaluation}
\label{app:evaluation_metrics}

On CounterFact and ZsRE, Efficacy measures success on edit prompts, Generalization on held-out paraphrases, and Specificity on unrelated queries. Utility is the arithmetic mean of these three metrics. Evaluation intervals are given in Appendix~\ref{app:editing_protocol}.

\paragraph{CounterFact.}
We compare the new and original answers using length-normalized log-likelihood. For prompt $x$, answer tokens $y=(y_1,\ldots,y_{|y|})$, and edited model $\mathcal{M}'$, the answer score is
\begin{equation}
    \operatorname{score}(x,y)
    = \frac{1}{|y|}\sum_{t=1}^{|y|}
    \log p_{\mathcal{M}'}(y_t\mid x,y_{<t}),
    \label{eq:eval_answer_score}
\end{equation}
where $y_{<t}$ contains the preceding reference-answer tokens. For edit case $i$, the edit prompt $x_i$ and held-out paraphrases succeed when the new answer $o_i^\star$ scores higher than the original answer $o_i$. Neighborhood prompts should retain the opposite preference. With $\mathbb{I}[\cdot]$ denoting the indicator function, the prompt-level scores are
\begin{equation}
    \begin{aligned}
        s_i^{\mathrm{edit}}(x)
        &= \mathbb{I}[\operatorname{score}(x,o_i^\star)>\operatorname{score}(x,o_i)],\\
        s_i^{\mathrm{nbr}}(x)
        &= \mathbb{I}[\operatorname{score}(x,o_i)>\operatorname{score}(x,o_i^\star)].
    \end{aligned}
    \label{eq:counterfact_prompt_scores}
\end{equation}
Efficacy averages $s_i^{\mathrm{edit}}(x_i)$ over edit cases. Generalization first averages $s_i^{\mathrm{edit}}(x)$ over each case's held-out paraphrases, then across cases. Specificity applies the same two-stage average to $s_i^{\mathrm{nbr}}(x)$ on neighborhood prompts. Scores are multiplied by 100, with each case weighted equally.

For generated continuations, Fluency averages $\frac{1}{3}H_2+\frac{2}{3}H_3$, where $H_2$ and $H_3$ are empirical word bigram and trigram entropies in bits. Consistency measures cosine similarity between TF-IDF vectors of the concatenated generated text and target-related reference text. We average scores across cases and multiply both metrics by 100 for reporting. Fluency is a scaled entropy, not a percentage accuracy.

\paragraph{ZsRE.}
Given the preceding reference-answer tokens, we check whether the model's most likely next token matches the next reference token. For prompt $x$ and reference answer $y$, this gives
\begin{equation}
    \operatorname{TokenAcc}(x,y)
    = \frac{1}{|y|}\sum_{t=1}^{|y|}
    \mathbb{I}\!\left[
        \operatorname*{arg\,max}_{v}
        p_{\mathcal{M}'}(v\mid x,y_{<t})=y_t
    \right],
    \label{eq:zsre_token_accuracy}
\end{equation}
where $v$ ranges over the token vocabulary. Efficacy evaluates the edit question and its target answer; Generalization uses the held-out paraphrase and the same answer. Specificity evaluates the unrelated question and its reference answer. Token accuracies are averaged within each case, then across cases, and multiplied by 100. Thus, Specificity measures post-edit answer accuracy, not agreement with pre-edit predictions. Newly correct and newly incorrect answers can offset each other; Appendix~\ref{app:zsre_correctness_transitions} separates these transitions.

\paragraph{MQuAKE.}
Standard prompting asks for a direct answer, while CoT prompting asks for reasoning before the final answer. Both use the benchmark's demonstrations and greedy decoding, with maximum generation lengths of 32 and 128 tokens, respectively. We extract the first nonempty answer line for standard prompting, removing an optional answer label, and the final \texttt{Answer:} line for CoT. Predictions and reference answers are normalized for capitalization, whitespace, and boundary punctuation. A prediction is correct if it exactly matches the revised answer or an accepted alias. A case is correct if at least one of its question variants is answered correctly. We report the percentage of correct cases.

\paragraph{General Capabilities.}
For each task, we select the candidate answer with the highest score in Eq.~\ref{eq:eval_answer_score} and compute class-frequency-weighted F1. For task $d$, let $\mathcal{C}_d$ be its label set, $n_{d,c}$ the number of examples with label $c$, and $n_d$ the total number of examples. The task score and the mean across six tasks are
\begin{equation}
    F_{1,d}
    = \sum_{c\in\mathcal{C}_d}\frac{n_{d,c}}{n_d}
    \frac{2\mathrm{TP}_{d,c}}
    {2\mathrm{TP}_{d,c}+\mathrm{FP}_{d,c}+\mathrm{FN}_{d,c}},
    \qquad
    \overline{F}_1=\frac{1}{6}\sum_{d=1}^{6}F_{1,d}.
    \label{eq:general_capability_f1}
\end{equation}
Here, $\mathrm{TP}_{d,c}$, $\mathrm{FP}_{d,c}$, and $\mathrm{FN}_{d,c}$ count true positives, false positives, and false negatives for class $c$; a zero denominator contributes zero. Classes are weighted by frequency within each task, and tasks are weighted equally. We measure changes relative to each method's own pre-edit performance. These F1 scores follow our evaluation protocol, not each benchmark's default leaderboard metric.

\paragraph{Confidence Intervals.}
Across the editing benchmarks, reported intervals quantify uncertainty across cases, not across random seeds. Following ROME~\citep{meng2022rome}, we use normal-approximation 95\% confidence intervals $\bar{x}\pm1.96s/\sqrt{N}$, where $\bar{x}$ and $s$ are the mean and standard deviation of case-level scores. Scores from multiple prompts within a case are averaged first. The sample size $N$ is the number of cases evaluated at each editing stage; the generated-expression analysis in Figure~\ref{fig:in_depth_analysis}(b) uses $N=2000$. MQuAKE instead uses Wilson 95\% confidence intervals over cases. Utility is reported as a point estimate.

\subsection{Baselines}
\label{app:baseline_methods}

We compare gradient-based expert editing, MoE-specific locate-then-edit, and direct conditional memory fine-tuning. AlphaEdit's original update rule applies to a dense FFN projection~\citep{fang2025alphaedit}. We exclude it because applying that rule to routed experts requires an additional choice of experts and update allocation~\citep{gu2026moeedit}. The gradient-based baselines use the model's existing routed computation, with editable parameters specified below.

\paragraph{FT and FT-L.}
FT uses gradient descent to update selected model parameters so that edit prompts produce the desired answers. FT-L additionally bounds each parameter change to limit deviation from the model entering the edit batch~\citep{zhu2020modifying}. In our MoE adaptation, both methods update expert FFN down-projections at layer 7. They use up to 25 epochs per edit batch, a learning rate of $5\times10^{-4}$, and optimization minibatches of one example. FT-L bounds each parameter change by $10^{-4}$.

\paragraph{AdaLoRA.}
AdaLoRA~\citep{zhang2023adalora} learns low-rank updates to weight matrices while keeping the pretrained weights fixed. It allocates more of the rank budget to important update components and prunes less important ones. We apply it to expert down-projections at layer 7, optimizing the prediction loss on edit prompts with initial rank 12, target rank 8, and scaling parameter 32. It uses 25 epochs per edit batch, a learning rate of $5\times10^{-4}$, and optimization minibatches of one example.

\paragraph{UnKE.}
UnKE~\citep{deng2025unke} edits knowledge by first computing target Transformer-block representations that lead to the desired answers. It then updates block parameters to match these targets while retaining the outputs on preservation examples. In our MoE adaptation, we fit the layer-7 block outputs by updating the experts' gate, up-, and down-projections; attention and router parameters remain fixed. Target computation uses up to 25 steps at learning rate $0.5$, and parameter fitting uses up to 50 steps at learning rate $2\times10^{-4}$.

\paragraph{MoEEdit.}
MoEEdit~\citep{gu2026moeedit} jointly updates expert FFN down-projections so that their router-weighted outputs match the editing targets. To preserve outputs on unrelated inputs and limit downstream routing shifts, it constrains each expert's update using a null-space projection built from preservation inputs. The updates are solved one expert at a time through block coordinate descent. We edit layer 7, using up to 25 target-computation steps at learning rate $0.1$ and four block coordinate descent passes. The per-expert null-space projections use statistics from 100,000 Wikipedia samples.

\paragraph{Memory-FT (Subject).}
This baseline directly fine-tunes conditional memory to predict the desired answers, restricting updates to embeddings activated at the last subject-token position of each original edit prompt. Since this selects a limited set of $n$-grams, we store one cumulative update vector per exact $n$-gram, as in EngramEdit. We optimize these vectors by backpropagating the prediction loss and add them whenever their $n$-grams are activated, including outside subject positions. The pretrained hashed tables, token embeddings, memory projection layers, and backbone remain fixed.

\paragraph{Memory-FT (All Tokens).}
This variant expands direct memory fine-tuning to embeddings activated at every token position in the original edit prompts, using the same prediction loss. Storing a separate update vector for each $n$-gram in this larger set would increase memory requirements. We therefore directly fine-tune the pretrained hashed tables, allowing rows activated at any token position to receive gradient updates while keeping all other parameters fixed. Both Memory-FT variants use $n$-gram lengths 2, 3, and 4, up to 25 epochs per edit batch, a learning rate of $3\times10^{-3}$, and optimization minibatches of one example.

\subsection{Implementation Details}
\label{app:implementation_details}

We describe the shared model and editing setup, followed by EngramEdit's target computation, memory mapping, and memory updating settings.

\subsubsection{Model and Editing Setup}
\label{app:editing_protocol}

All experiments use LongCat-Flash-Lite, a 68.5B-parameter MoE model with 31.4B parameters in $n$-gram embeddings~\citep{liu2026scalingembeddings}, and run on two NVIDIA RTX PRO 6000 GPUs. Its conditional memory uses suffix $n$-grams of lengths 2, 3, and 4, with four hashed sub-tables per length. Each sub-table returns a 256-dimensional vector, which is linearly projected to the 3,072-dimensional token embedding space. At each position, the model averages the twelve projected vectors and the original token embedding before passing the result to the backbone. EngramEdit keeps the token embeddings and these projection matrices fixed.

\paragraph{Sequential Editing.}
An edit batch contains the requests applied together; an optimization minibatch contains the examples used in one gradient step. The main CounterFact and ZsRE experiments apply 2,000 edits in batches of 100, with each batch starting from the model state produced by preceding batches. The longer runs extend editing to 5,000 facts. For MQuAKE, we shuffle the 3,000 cases once with seed 0 and use the same case IDs and hop labels for all methods. Each batch contains 100 cases, with all their fact-level edit requests applied together, so a batch can contain more than 100 fact edits.

\paragraph{Evaluation Schedule.}
CounterFact trajectories are evaluated every 100 edits, and ZsRE trajectories are compared at common intervals of 200 edits. General capabilities are evaluated before editing and every 500 edits through 5,000 sequential CounterFact edits, using 100 examples per task (see Appendix~\ref{app:datasets}).

\subsubsection{Target Computation and Memory Mapping}
\label{app:implementation_target_computation}

For each edit request, we ask the model to generate four semantically equivalent expressions before editing, using the prompt and filtering rules in Appendix~\ref{app:expression_generation}. The retained expressions and original edit prompt form the expression set used for target computation and memory mapping.

\paragraph{Prediction Loss.}
During target computation, we evaluate each expression both alone and after each of five model-generated prefix contexts, which vary the surrounding text without rephrasing the edit request. The shared perturbation is added to the conditional memory representation before the first decoder block, while all decoder parameters remain fixed. Following the target readout in MEMIT~\citep{meng2023memit}, we compute the NLL from the last decoder block's hidden states using the model's fixed final normalization and language-model head. This yields the final predictive distribution in Eq.~\ref{eq:appendix_target_nll}; the KL preservation term also uses final-output probabilities.

\paragraph{Optimization Settings.}
We optimize the shared perturbation with Adam for at most 25 steps at learning rate $0.5$, with $\lambda_{\mathrm{KL}}=0.0625$, $\lambda_{\mathrm{norm}}=0.001$, and $\rho=32$. Optimization stops when the total target-computation loss falls below $0.05$, checked every two steps.

\paragraph{Memory Mapping.}
We use expressions without prefix contexts for memory mapping. At each expression's last subject-token position, we extract suffix $n$-grams of lengths 2, 3, and 4, skipping incomplete windows and windows crossing an end-of-sequence token.

\subsubsection{Memory Updating}
\label{app:implementation_memory_updating}

Conditional memory uses hashed tables to control the storage cost of a large $n$-gram space. Editing only requires storing updates for the selected $n$-grams. We therefore maintain a separate cumulative update vector for each edited $n$-gram, indexed by its exact token sequence. Additional storage scales with the number of distinct edited $n$-grams, and the exact indexing prevents hash collisions from coupling their updates. Each vector is added to the embedding output whenever its $n$-gram is activated, while the pretrained tables remain fixed.

\paragraph{Regularization and Solver Settings.}
We use the weight construction in Eq.~\ref{eq:reuse_weight}, setting $\lambda_{\mathrm{ridge}}=0.01$ and $\lambda_{\mathrm{reuse}}=0.05$. The length weights are $\eta_2=8$, $\eta_3=2$, and $\eta_4=1$. For frequency-based regularization, we use $\gamma=9$, $\beta=1$, $w_{\mathrm{freq}}^{\max}=10$, and $w^{\max}=64$. We estimate the corpus frequency of the selected $n$-grams from up to 3,000,000 documents in the November 1, 2023 English Wikipedia snapshot, using at most 4,096 tokens from each document. Within each $n$-gram length, the percentile $\pi_j$ is the normalized rank of the observed count among the distinct count values. Missing $n$-grams receive a frequency weight of 1. We form and solve the joint linear system in FP32 using a direct linear solver.

\section{Additional Experimental Results}
\label{app:additional_results}

This section provides detailed evaluation results and analyses of editing choices and memory use to complement Section~\ref{sec:experiments}.

\subsection{Detailed Evaluation Results}
\label{app:detailed_evaluation_results}

We extend the sequential editing results, examine multi-hop reasoning by depth, and analyze ZsRE Specificity to complement Section~\ref{sec:overall_performance}. We then report task-level general-capability trajectories to complement Section~\ref{sec:general_capabilities}.

\subsubsection{Sequential Editing with 5,000 Facts}
\label{app:extended_editing_results}

\begin{figure}[t]
    \centering
    \includegraphics[width=\linewidth]{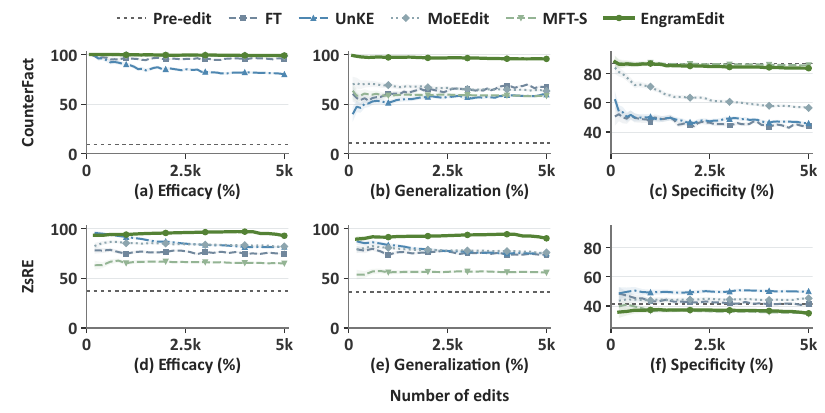}
    \caption{Sequential editing on CounterFact (a--c) and ZsRE (d--f) up to 5,000 edits. Shading shows pointwise case-level 95\% confidence intervals. MFT-S denotes Memory-FT (Subject).}

    \label{fig:extended_editing_results}
\end{figure}

To test whether knowledge updates through conditional memory remain effective as edits accumulate, we extend the CounterFact and ZsRE experiments in Section~\ref{sec:overall_performance} to 5,000 sequential edits in batches of 100. We compare EngramEdit with FT, UnKE, MoEEdit, and MFT-S. Each evaluation covers all edits applied so far. Figure~\ref{fig:extended_editing_results} reports Efficacy, Generalization, and Specificity every 100 edits on CounterFact and every 200 edits on ZsRE. We observe that:

\begin{itemize}[leftmargin=1em, labelsep=0.4em, nosep]
\item EngramEdit maintains high Efficacy and a clear Generalization lead on both datasets through 5,000 edits. Because each evaluation includes all earlier edits, these results show that revised knowledge remains accessible across expressions as further updates accumulate. EngramEdit supports this access by jointly updating embeddings across expressions, allowing different wordings to activate memory updated for the same fact.
\item On CounterFact, EngramEdit and MFT-S maintain much higher Specificity than FT, UnKE, and MoEEdit throughout the sequence, but EngramEdit achieves substantially higher Generalization at similar Efficacy. EngramEdit therefore combines cross-expression access with largely preserved unrelated knowledge as edits accumulate. Reuse-based regularization can help explain this preservation because it penalizes large updates to embeddings that unrelated inputs are more likely to activate.
\end{itemize}

\subsubsection{Multi-Hop Reasoning}
\label{app:mquake_hop_results}

To test whether knowledge updated through conditional memory supports reasoning at different depths, we break down the MQuAKE evaluation in Section~\ref{sec:overall_performance} by hop count. All eight methods use the same 3,000 cases: 1,135 two-hop, 1,136 three-hop, and 729 four-hop cases. We compare standard prompting, which asks for a direct answer, with CoT prompting, which asks for step-by-step reasoning before answering. Table~\ref{tab:mquake_hop_results} reports accuracy within each group. We observe that:

\begin{table}[t]
    \centering
    \caption{MQuAKE accuracy (\%) by hop with standard and CoT prompts. Brackets give Wilson 95\% confidence intervals. Bold and underlining mark the best and second-best scores.}
    \label{tab:mquake_hop_results}
    \begingroup
    \small
    \setlength{\tabcolsep}{5pt}
    \renewcommand{\arraystretch}{1}
    \newcommand{\hopci}[3]{\shortstack{#1\\{\fontsize{6.5}{7.5}\selectfont [#2, #3]}}}
    \begin{tabular*}{\linewidth}{@{\extracolsep{\fill}}lcccccc@{}}
        \toprule
        \multirow{2}{*}[-0.5ex]{\textbf{Method}} & \multicolumn{3}{c}{\textbf{Standard}\,$\boldsymbol{\uparrow}$} & \multicolumn{3}{c}{\textbf{CoT}\,$\boldsymbol{\uparrow}$} \\
        \cmidrule(lr){2-4}\cmidrule(lr){5-7}
        & \textbf{2-hop} & \textbf{3-hop} & \textbf{4-hop} & \textbf{2-hop} & \textbf{3-hop} & \textbf{4-hop} \\
        \midrule
        FT & \hopci{\underline{7.93}}{6.50}{9.65} & \hopci{2.11}{1.42}{3.12} & \hopci{\textbf{7.82}}{6.08}{10.00} & \hopci{\underline{12.60}}{10.79}{14.66} & \hopci{3.61}{2.67}{4.86} & \hopci{\underline{9.74}}{7.79}{12.11} \\
        FT-L & \hopci{1.59}{1.01}{2.49} & \hopci{1.67}{1.07}{2.60} & \hopci{4.53}{3.24}{6.29} & \hopci{1.06}{0.61}{1.84} & \hopci{0.97}{0.54}{1.73} & \hopci{1.78}{1.05}{3.03} \\
        AdaLoRA & \hopci{1.50}{0.94}{2.39} & \hopci{2.64}{1.86}{3.74} & \hopci{3.84}{2.67}{5.50} & \hopci{1.15}{0.67}{1.95} & \hopci{1.23}{0.74}{2.06} & \hopci{2.33}{1.46}{3.70} \\
        UnKE & \hopci{4.14}{3.13}{5.46} & \hopci{\textbf{4.23}}{3.20}{5.56} & \hopci{\underline{6.17}}{4.65}{8.16} & \hopci{5.46}{4.28}{6.94} & \hopci{\underline{4.58}}{3.51}{5.95} & \hopci{9.60}{7.67}{11.96} \\
        MoEEdit & \hopci{3.00}{2.15}{4.16} & \hopci{2.64}{1.86}{3.74} & \hopci{3.29}{2.22}{4.85} & \hopci{0.62}{0.30}{1.27} & \hopci{0.26}{0.09}{0.77} & \hopci{0.27}{0.08}{0.99} \\
        MFT-S & \hopci{6.61}{5.30}{8.20} & \hopci{2.55}{1.78}{3.64} & \hopci{5.49}{4.06}{7.39} & \hopci{8.28}{6.82}{10.03} & \hopci{3.61}{2.67}{4.86} & \hopci{3.98}{2.78}{5.65} \\
        MFT-A & \hopci{5.02}{3.90}{6.45} & \hopci{1.32}{0.80}{2.17} & \hopci{5.62}{4.17}{7.54} & \hopci{4.32}{3.28}{5.66} & \hopci{2.02}{1.35}{3.02} & \hopci{3.57}{2.45}{5.17} \\
        EngramEdit & \hopci{\textbf{9.52}}{7.94}{11.36} & \hopci{\underline{2.82}}{2.00}{3.95} & \hopci{5.49}{4.06}{7.39} & \hopci{\textbf{36.92}}{34.16}{39.76} & \hopci{\textbf{20.07}}{17.84}{22.50} & \hopci{\textbf{15.09}}{12.67}{17.87} \\
        \bottomrule
    \end{tabular*}
    \endgroup
\end{table}

\begin{itemize}[leftmargin=1em, labelsep=0.4em, nosep]
\item EngramEdit gains substantially from CoT at all three depths, whereas FT-L, AdaLoRA, and MoEEdit lose accuracy in every group. This contrast shows that the benefit of CoT depends on the editing method, not just the reasoning prompt. For EngramEdit, CoT can help because intermediate steps explicitly express facts whose $n$-grams activate updated embeddings, allowing revised information to guide subsequent reasoning.
\item The best method varies across hop groups under standard prompting, but EngramEdit leads every group with CoT. Its advantage persists through four-hop questions, despite lower accuracy and smaller margins at greater depths. This supports the use of revised knowledge in longer reasoning chains, which require more facts to be accessed and combined.
\end{itemize}

\subsubsection{Analysis of ZsRE Specificity}
\label{app:zsre_correctness_transitions}

To understand whether higher ZsRE Specificity reflects better preservation, we compare correctness on unrelated queries before and after 2,000 edits. All seven methods are evaluated on the same 11,379 neighborhood prompts from the first 2,000 edit cases. We compare their recorded post-edit correctness labels against one shared evaluation of unedited LongCat. EngramEdit's results come from the first 2,000 edits of the 5,000-edit run, separate from the main-table run.

Let C and W denote correct and wrong predictions. For each transition $a\!\rightarrow\!b$, we compute its proportion over all prompts within each case, then average equally across cases to obtain $P(a\!\rightarrow\!b)$. Post-edit Specificity counts both retained correct predictions ($\mathrm{C}\!\rightarrow\!\mathrm{C}$) and newly correct predictions ($\mathrm{W}\!\rightarrow\!\mathrm{C}$). We report the change in Specificity and Stable Correctness, the proportion of unchanged correctness states:
\begin{equation}
    \begin{aligned}
        \Delta\text{Specificity} &= P(\mathrm{W}\!\rightarrow\!\mathrm{C})-P(\mathrm{C}\!\rightarrow\!\mathrm{W}),\\
        \text{Stable Correctness} &= 1-P(\mathrm{C}\!\rightarrow\!\mathrm{W})-P(\mathrm{W}\!\rightarrow\!\mathrm{C}).
    \end{aligned}
    \label{eq:zsre_correctness_transitions}
\end{equation}
Stable Correctness includes predictions that remain wrong, so it differs from retention of previously correct predictions. It also does not imply identical outputs, since one wrong token can change to another. We report 95\% confidence-interval half-widths of $1.95996s/\sqrt{2000}$, where $s$ is the sample standard deviation of the case-level proportions. Table~\ref{tab:zsre_correctness_transitions} summarizes the results. We observe that:

\begin{table}[t]
    \centering
    \caption{ZsRE correctness transitions after 2,000 edits. C and W denote correct and wrong predictions; Post--Pre is the Specificity change in percentage points. Stable Corr. measures unchanged correctness states. Values are case-averaged percentages with 95\% CI half-widths in subscripts.}
    \label{tab:zsre_correctness_transitions}
    \begingroup
    \fontsize{9}{10}\selectfont
    \setlength{\tabcolsep}{5pt}
    \renewcommand{\arraystretch}{1}
    \newcommand{\tabci}[2]{\ensuremath{\text{#1}{}_{\scriptscriptstyle\pm\text{#2}}}}
    \begin{tabular*}{\linewidth}{@{\extracolsep{\fill}}lccccc@{}}
        \toprule
        \textbf{Method} & $\boldsymbol{\mathrm{C}\!\rightarrow\!\mathrm{W}}$ & $\boldsymbol{\mathrm{W}\!\rightarrow\!\mathrm{C}}$ & \textbf{Post--Pre} & \textbf{Post Spec.}\,$\boldsymbol{\uparrow}$ & \textbf{Stable Corr.}\,$\boldsymbol{\uparrow}$ \\
        \midrule
        FT-L & \tabci{0.62}{0.19} & \tabci{4.84}{0.57} & +4.23 & \tabci{45.76}{1.24} & \tabci{94.54}{0.60} \\
        AdaLoRA & \tabci{2.25}{0.33} & \tabci{7.51}{0.71} & +5.26 & \tabci{46.79}{1.27} & \tabci{90.24}{0.79} \\
        UnKE & \tabci{2.95}{0.36} & \tabci{11.26}{0.86} & +8.31 & \tabci{49.85}{1.32} & \tabci{85.79}{0.93} \\
        MoEEdit & \tabci{2.30}{0.35} & \tabci{5.14}{0.56} & +2.84 & \tabci{44.37}{1.26} & \tabci{92.56}{0.67} \\
        MFT-S & \tabci{5.06}{0.51} & \tabci{0.40}{0.12} & -4.67 & \tabci{36.87}{1.17} & \tabci{94.54}{0.54} \\
        MFT-A & \tabci{9.67}{0.72} & \tabci{1.84}{0.30} & -7.83 & \tabci{33.71}{1.13} & \tabci{88.48}{0.80} \\
        EngramEdit & \tabci{5.25}{0.54} & \tabci{0.92}{0.25} & -4.33 & \tabci{37.21}{1.18} & \tabci{93.83}{0.60} \\
        \bottomrule
    \end{tabular*}
    \endgroup
\end{table}

\begin{itemize}[leftmargin=1em, labelsep=0.4em, nosep]
\item FT-L, AdaLoRA, UnKE, and MoEEdit exceed pre-edit Specificity because their $\mathrm{W}\!\rightarrow\!\mathrm{C}$ rates exceed their $\mathrm{C}\!\rightarrow\!\mathrm{W}$ rates. Their higher post-edit accuracy than EngramEdit reflects both better retention of initially correct predictions and more corrections of initially wrong predictions.
\item EngramEdit preserves 93.83\% of correctness states and about 87.4\% of initially correct predictions under case-weighted aggregation. Its $\mathrm{W}\!\rightarrow\!\mathrm{C}$ rate of only 0.92\% shows that its post-edit Specificity comes predominantly from retained correct predictions. Together with the high Efficacy and Generalization reported in Section~\ref{sec:overall_performance}, these results show that EngramEdit supports effective, generalizable knowledge updates while retaining most previously correct predictions on unrelated queries.
\end{itemize}

\subsubsection{Task-Level General Capabilities}
\label{app:task_level_capabilities}

To assess capability preservation across individual tasks as edits accumulate, we track six task-level F1 scores during 5,000 sequential CounterFact edits in batches of 100. We compare EngramEdit with FT, UnKE, MoEEdit, and MFT-S on 100 examples per task, evaluating before editing and every 500 edits. Figure~\ref{fig:task_level_capabilities} shows each method's class-frequency-weighted F1 trajectory, with pre-edit scores as references. We observe that:

\begin{figure}[t]
    \centering
    \includegraphics[width=\linewidth]{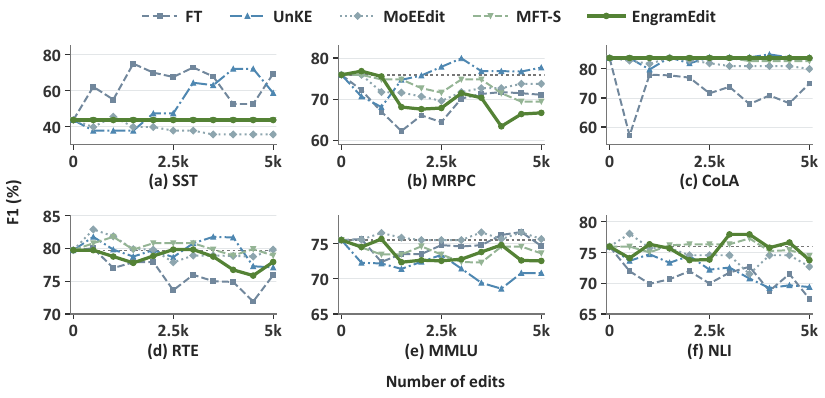}
    \caption{Task-level general capabilities during 5,000 sequential CounterFact edits. F1 is measured on 100 examples per task. Dotted lines mark pre-edit scores.}
    \label{fig:task_level_capabilities}
\end{figure}

\begin{itemize}[leftmargin=1em, labelsep=0.4em, nosep]
\item EngramEdit preserves pre-edit F1 on SST and CoLA throughout editing, while RTE, MMLU, and NLI stay within four F1 points of their pre-edit scores. By comparison, FT and UnKE show large SST gains and NLI losses. These trajectories show that EngramEdit largely preserves performance across five tasks throughout sequential editing. This stability can be partly explained by restricting edits to selected embeddings and applying stronger penalties to frequently reused ones, which limits changes to memory shared with unrelated inputs.
\item EngramEdit's larger changes are concentrated on MRPC, and MFT-S shows a similar pattern. This suggests that MRPC is more sensitive to memory updates than the other evaluated tasks. MRPC requires comparing the meanings of two sentences, so updates that affect their representations differently may change the equivalence judgment. Such differences can arise when the sentences activate different $n$-grams.
\end{itemize}

\subsection{Analysis of Editing Choices}
\label{app:additional_ablation_studies}

To examine how editing choices affect knowledge updating and preservation, we first compare editors with and without generated expressions. We then vary EngramEdit's expression count, editable $n$-gram lengths, regularization coefficients, and clamp factor. These studies complement the component analyses in Section~\ref{sec:in_depth_analysis}.

\subsubsection{Multi-Expression Editing}
\label{app:multi_expression_editing}

To examine how generated expressions affect different editors, we compare MoEEdit, MFT-S, MFT-A, and EngramEdit with zero or four generated expressions per fact on CounterFact and ZsRE. Each setting uses 2,000 sequential edits in batches of 100 and retains the original edit prompt. All methods use the same generated expressions in the four-expression setting. From Table~\ref{tab:multi_expression_editing}, we observe that:

\begin{table}[t]
    \centering
    \caption{Editing with and without generated expressions on CounterFact and ZsRE over 2,000 sequential edits. Subscripts give 95\% CI half-widths. Bold and underlining mark the best and second-best scores within each block.}
    \label{tab:multi_expression_editing}
    \begingroup
    \fontsize{9}{10}\selectfont
    \setlength{\tabcolsep}{1pt}
    \renewcommand{\arraystretch}{1.12}
    \newcommand{\tabci}[2]{\ensuremath{\text{#1}{}_{\scriptscriptstyle\pm\text{#2}}}}
    \newcommand{\metrichead}[1]{\textbf{#1}\,$\boldsymbol{\uparrow}$}
    \begin{tabular*}{\linewidth}{@{\extracolsep{\fill}}lcccccccc@{}}
        \toprule
        \multirow{2}{*}[-0.5ex]{\textbf{Method}} & \multicolumn{4}{c}{\textbf{CounterFact}} & \multicolumn{4}{c}{\textbf{ZsRE}} \\
        \cmidrule(lr){2-5}\cmidrule(lr){6-9}
        & \metrichead{Eff.} & \metrichead{Gen.} & \metrichead{Spec.} & \metrichead{Util.} & \metrichead{Eff.} & \metrichead{Gen.} & \metrichead{Spec.} & \metrichead{Util.} \\
        \midrule
        \rowcolor{black!6}
        \multicolumn{9}{l}{\textbf{Original edit prompt only}} \\
        MoEEdit & \tabci{\underline{99.1}}{0.41} & \tabci{\underline{67.1}}{1.69} & \tabci{63.5}{1.28} & 76.6 & \tabci{\underline{85.1}}{0.91} & \tabci{\underline{78.7}}{1.21} & \tabci{\textbf{44.4}}{1.26} & \underline{69.4} \\
        MFT-S & \tabci{99.0}{0.45} & \tabci{59.1}{1.92} & \tabci{\underline{85.2}}{0.90} & \underline{81.1} & \tabci{66.9}{1.35} & \tabci{56.5}{1.53} & \tabci{36.9}{1.17} & 53.4 \\
        MFT-A & \tabci{93.9}{1.04} & \tabci{40.7}{1.75} & \tabci{66.6}{1.12} & 67.1 & \tabci{62.8}{1.49} & \tabci{51.3}{1.57} & \tabci{33.7}{1.13} & 49.3 \\
        EngramEdit & \tabci{\textbf{99.5}}{0.32} & \tabci{\textbf{75.0}}{1.74} & \tabci{\textbf{85.6}}{0.91} & \textbf{86.7} & \tabci{\textbf{95.1}}{0.62} & \tabci{\textbf{83.4}}{1.29} & \tabci{\underline{38.5}}{1.18} & \textbf{72.3} \\
        \addlinespace[2pt]
        \rowcolor{black!6}
        \multicolumn{9}{l}{\textbf{Original edit prompt + 4 generated expressions}} \\
        MoEEdit & \tabci{80.7}{1.73} & \tabci{59.8}{1.71} & \tabci{66.0}{1.21} & 68.8 & \tabci{\underline{79.8}}{1.12} & \tabci{\underline{72.9}}{1.35} & \tabci{\textbf{42.8}}{1.23} & \underline{65.2} \\
        MFT-S & \tabci{\underline{99.0}}{0.45} & \tabci{\underline{90.7}}{1.09} & \tabci{\underline{85.0}}{0.91} & \underline{91.5} & \tabci{67.2}{1.37} & \tabci{64.2}{1.43} & \tabci{37.7}{1.17} & 56.4 \\
        MFT-A & \tabci{93.8}{1.06} & \tabci{63.2}{1.72} & \tabci{63.6}{1.20} & 73.5 & \tabci{69.8}{1.31} & \tabci{65.9}{1.43} & \tabci{31.9}{1.11} & 55.9 \\
        EngramEdit & \tabci{\textbf{99.5}}{0.31} & \tabci{\textbf{97.0}}{0.69} & \tabci{\textbf{85.2}}{0.92} & \textbf{93.9} & \tabci{\textbf{97.3}}{0.40} & \tabci{\textbf{93.7}}{0.79} & \tabci{\underline{38.3}}{1.18} & \textbf{76.4} \\
        \bottomrule
    \end{tabular*}
    \endgroup
\end{table}

\begin{itemize}[leftmargin=1em, labelsep=0.4em, nosep]
\item Adding four generated expressions improves Generalization for MFT-S, MFT-A, and EngramEdit on both datasets, whereas MoEEdit's Efficacy and Generalization decrease. The benefit therefore depends on how each editor uses the expressions. Generated expressions can help memory-only editors because their different wordings activate additional $n$-grams, allowing held-out paraphrases to access revised facts through a larger set of updated embeddings.
\item EngramEdit leads in Efficacy, Generalization, and Utility on both datasets with either expression count. On CounterFact, it generalizes better than MFT-S at similar Efficacy and Specificity when both use the same expressions and memory update interface. On ZsRE, its stronger Efficacy and Generalization yield higher Utility than MoEEdit despite lower Specificity. EngramEdit thus uses the same expressions more effectively for knowledge updating. By computing targets across expressions and matching them jointly, it accounts for the requirements of all expressions sharing each embedding. Reuse-based regularization further limits large changes to embeddings also used by unrelated inputs.
\end{itemize}

\subsubsection{Number of Generated Expressions}
\label{app:expression_count}

To assess the benefit of generating more expressions per edit, we compare $K\in\{0,2,4,6,8\}$ generated expressions on CounterFact and ZsRE. Each setting uses 2,000 sequential edits in batches of 100. The original edit prompt is always retained, and all other settings are fixed. From Table~\ref{tab:expression_count}, we observe that:

\begin{table}[t]
    \centering
    \caption{Results with different numbers of generated expressions after 2,000 edits. \# Expr. excludes the original prompt. Subscripts give reported CI half-widths; bold and underlining mark best and second-best scores.}
    \label{tab:expression_count}
    \begingroup
    \fontsize{9}{10}\selectfont
    \setlength{\tabcolsep}{1pt}
    \renewcommand{\arraystretch}{1.12}
    \newcommand{\tabci}[2]{\ensuremath{\text{#1}{}_{\scriptscriptstyle\pm\text{#2}}}}
    \newcommand{\metrichead}[1]{\textbf{#1}\,$\boldsymbol{\uparrow}$}
    \begin{tabular*}{\linewidth}{@{\extracolsep{\fill}}ccccccccccc@{}}
        \toprule
        \multirow{2}{*}[-0.5ex]{\textbf{\# Expr.}} & \multicolumn{6}{c}{\textbf{CounterFact}} & \multicolumn{4}{c}{\textbf{ZsRE}} \\
        \cmidrule(lr){2-7}\cmidrule(lr){8-11}
        & \metrichead{Eff.} & \metrichead{Gen.} & \metrichead{Spec.} & \metrichead{Util.} & \metrichead{Flu.} & \metrichead{Cons.} & \metrichead{Eff.} & \metrichead{Gen.} & \metrichead{Spec.} & \metrichead{Util.} \\
        \midrule
        0 & \tabci{\textbf{99.5}}{0.32} & \tabci{75.0}{1.74} & \tabci{\textbf{85.6}}{0.91} & 86.7 & \tabci{562.6}{1.19} & \tabci{15.7}{0.36} & \tabci{95.1}{0.62} & \tabci{83.4}{1.29} & \tabci{\textbf{38.5}}{1.18} & 72.3 \\
        2 & \tabci{\underline{99.4}}{0.34} & \tabci{95.2}{0.83} & \tabci{\underline{85.3}}{0.92} & 93.3 & \tabci{\underline{566.2}}{1.11} & \tabci{\textbf{16.2}}{0.36} & \tabci{96.2}{0.50} & \tabci{91.8}{0.89} & \tabci{37.7}{1.10} & 75.2 \\
        4 & \tabci{\textbf{99.5}}{0.31} & \tabci{97.0}{0.69} & \tabci{85.2}{0.92} & \underline{93.9} & \tabci{565.2}{1.10} & \tabci{\underline{16.1}}{0.36} & \tabci{\textbf{97.3}}{0.40} & \tabci{\underline{93.7}}{0.79} & \tabci{\underline{38.3}}{1.18} & \textbf{76.4} \\
        6 & \tabci{\underline{99.4}}{0.34} & \tabci{\textbf{97.3}}{0.64} & \tabci{85.2}{0.92} & \textbf{94.0} & \tabci{\textbf{567.5}}{1.03} & \tabci{\textbf{16.2}}{0.36} & \tabci{96.5}{0.47} & \tabci{\underline{93.7}}{0.77} & \tabci{36.8}{1.17} & 75.7 \\
        8 & \tabci{99.1}{0.41} & \tabci{\underline{97.2}}{0.67} & \tabci{84.9}{0.93} & 93.7 & \tabci{566.0}{1.07} & \tabci{16.0}{0.36} & \tabci{\underline{96.8}}{0.44} & \tabci{\textbf{94.2}}{0.74} & \tabci{38.0}{1.17} & \underline{76.3} \\
        \bottomrule
    \end{tabular*}
    \endgroup
\end{table}

\begin{itemize}[leftmargin=1em, labelsep=0.4em, nosep]
\item Two generated expressions provide most of the Generalization gain on both datasets, with further improvements at four. Efficacy remains high, while Specificity on both datasets and Fluency and Consistency on CounterFact vary little. A small expression set thus supports broader access to revised knowledge with largely preserved unrelated-query performance. Generated expressions enable this access by adding the $n$-grams activated by different wordings to memory mapping, so held-out paraphrases have more opportunities to activate updated embeddings.
\item Beyond four expressions, Generalization improves only slightly and Utility does not consistently increase. Four expressions achieve the highest Utility on ZsRE and nearly the highest on CounterFact. This supports $K=4$ as the default, since larger sets require more expression processing without comparable performance gains.
\end{itemize}

\subsubsection{Editable \texorpdfstring{$n$}{n}-gram Lengths}
\label{app:editable_ngram_lengths}

To identify which $n$-gram lengths support both effective updates and preservation, we compare eight editable-length configurations on CounterFact using 2,000 sequential edits in batches of 100. We test lengths 2, 3, and 4 individually, in pairs, and together, as well as the full combination with single-token ($n=1$) embedding updates. All settings use four generated expressions and reuse-based regularization; only the lengths eligible for embedding updates change. From Figure~\ref{fig:editable_ngram_lengths}, we observe that:

\begin{figure}[t]
    \centering
    \includegraphics[width=\linewidth]{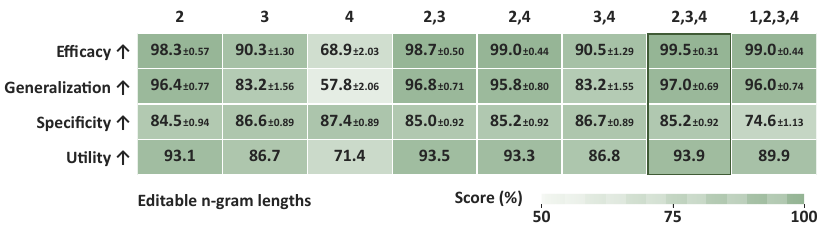}
    \caption{Editable $n$-gram lengths on CounterFact. The frame marks the default lengths 2, 3, and 4; $n=1$ denotes single-token embedding updates. Small $\pm$ values give CI half-widths.}
    \label{fig:editable_ngram_lengths}
\end{figure}

\begin{itemize}[leftmargin=1em, labelsep=0.4em, nosep]
\item Among single-length settings, 2-grams achieve the highest Efficacy and Generalization, while longer $n$-grams give higher Specificity. Shorter $n$-grams therefore favor access across expressions, whereas longer ones favor preservation. This is because shorter token sequences can activate the same updated embedding across both paraphrases and unrelated inputs.
\item Combining lengths 2, 3, and 4 achieves the highest Efficacy, Generalization, and Utility, with higher Specificity than using 2-grams alone. This suggests that different lengths can complement each other. Shorter $n$-grams support access across expressions, while longer ones provide more context-specific embeddings that can carry part of the update, reducing reliance on embeddings shared with unrelated inputs.
\item Excluding single-token embedding updates yields markedly higher Specificity than editing lengths 1--4, with no loss in Efficacy or Generalization. This supports restricting the default editable lengths to 2--4. Individual tokens occur in many unrelated inputs, so single-token updates can affect a wider range of predictions without providing additional editing gains here.
\end{itemize}

\begin{table}[t]
    \centering
    \caption{Sensitivity to regularization and clamp factor after 2,000 CounterFact edits. Only the listed parameter changes. Scores are percentages; subscripts give reported CI half-widths. Bold parameter values mark the defaults.}
    \label{tab:regularization_clamp}
    \begingroup
    \fontsize{9}{10}\selectfont
    \setlength{\tabcolsep}{5pt}
    \renewcommand{\arraystretch}{1.08}
    \newcommand{\tabci}[2]{\ensuremath{\text{#1}{}_{\scriptscriptstyle\pm\text{#2}}}}
    \newcommand{\metrichead}[1]{\textbf{#1}\,$\boldsymbol{\uparrow}$}
    \begin{tabular*}{\linewidth}{@{\extracolsep{\fill}}lccccc@{}}
        \toprule
        \textbf{Parameter} & \textbf{Value} & \metrichead{Eff.} & \metrichead{Gen.} & \metrichead{Spec.} & \metrichead{Util.} \\
        \midrule
        \multirow{5}{*}{\shortstack[l]{Reuse-based regularization\\$\lambda_{\mathrm{reuse}}$}} & 0 & \tabci{98.8}{0.49} & \tabci{96.5}{0.76} & \tabci{84.7}{0.93} & 93.3 \\
        & 0.03 & \tabci{99.3}{0.37} & \tabci{96.2}{0.76} & \tabci{85.2}{0.92} & 93.6 \\
        & \textbf{0.05} & \tabci{99.5}{0.31} & \tabci{97.0}{0.69} & \tabci{85.2}{0.92} & 93.9 \\
        & 0.075 & \tabci{99.6}{0.29} & \tabci{96.4}{0.73} & \tabci{85.3}{0.91} & 93.8 \\
        & 0.1 & \tabci{99.6}{0.28} & \tabci{96.6}{0.70} & \tabci{85.2}{0.92} & 93.8 \\
        \midrule
        \multirow{5}{*}{\shortstack[l]{Ridge regularization\\$\lambda_{\mathrm{ridge}}$}} & 0 & \tabci{99.5}{0.31} & \tabci{96.6}{0.71} & \tabci{85.1}{0.92} & 93.7 \\
        & 0.001 & \tabci{99.6}{0.29} & \tabci{96.6}{0.72} & \tabci{85.3}{0.92} & 93.8 \\
        & \textbf{0.01} & \tabci{99.5}{0.31} & \tabci{97.0}{0.69} & \tabci{85.2}{0.92} & 93.9 \\
        & 0.03 & \tabci{99.4}{0.35} & \tabci{96.8}{0.71} & \tabci{85.3}{0.92} & 93.8 \\
        & 0.1 & \tabci{99.1}{0.41} & \tabci{96.6}{0.73} & \tabci{84.8}{0.93} & 93.5 \\
        \midrule
        \multirow{4}{*}{Clamp factor $\rho$} & 8 & \tabci{99.1}{0.41} & \tabci{97.0}{0.69} & \tabci{85.2}{0.92} & 93.8 \\
        & 16 & \tabci{99.4}{0.35} & \tabci{96.7}{0.71} & \tabci{85.3}{0.92} & 93.8 \\
        & \textbf{32} & \tabci{99.5}{0.31} & \tabci{97.0}{0.69} & \tabci{85.2}{0.92} & 93.9 \\
        & 64 & \tabci{99.6}{0.28} & \tabci{96.4}{0.73} & \tabci{85.0}{0.93} & 93.7 \\
        \bottomrule
    \end{tabular*}
    \endgroup
\end{table}

\subsubsection{Regularization and Clamp Factor}
\label{app:regularization_clamp}

To test whether strong editing performance requires narrowly tuned regularization or perturbation bounds, we vary one parameter at a time on CounterFact using 2,000 sequential edits in batches of 100. The reuse coefficient $\lambda_{\mathrm{reuse}}$ scales the length- and frequency-based penalties, while the ridge coefficient $\lambda_{\mathrm{ridge}}$ penalizes all embedding updates equally. The clamp factor $\rho$ bounds the perturbation norm during target computation. Table~\ref{tab:regularization_clamp} lists the tested values and marks the defaults. Other settings remain fixed, including four generated expressions and editable lengths 2, 3, and 4. We observe that:

\begin{itemize}[leftmargin=1em, labelsep=0.4em, nosep]
\item All tested nonzero reuse coefficients yield higher Efficacy, Specificity, and Utility than disabling the reuse-based term, with little variation among them. The gains therefore do not depend on one precisely tuned strength. Changing the coefficient scales the reuse-based penalties but retains stronger constraints on frequently reused embeddings, which can help preserve unrelated knowledge across these settings.
\item All four metrics remain close across ridge coefficients, including zero. This shows that strong performance does not require a precisely tuned uniform penalty. Reuse-based regularization still constrains embedding updates when ridge is removed, so the extra term mainly provides another way to control overall update size.
\item Efficacy, Generalization, and Specificity remain stable as the clamp factor increases from 8 to 64, with no consistent Utility gain from larger values. Strong editing performance thus holds across the tested norm bounds. A larger factor permits, but does not force, a larger perturbation, since prediction loss and target regularization still guide its optimization.
\end{itemize}

\subsection{Analysis of Memory Use}
\label{app:additional_memory_analysis}

To understand how revised facts are accessed through conditional memory, we first test whether successful recall depends on fact-related updates by selectively disabling them, extending Section~\ref{sec:memory_update_analysis}. With all updates enabled, we next group inputs by the updated embeddings they activate, then analyze how cross-edit sharing relates to editing outcomes as updates accumulate. These later analyses leave the edited model unchanged and identify associations, while disabling directly tests the effect of removing updates.

\subsubsection{Disabling Memory Updates}
\label{app:memory_disabling}

\begin{figure}[t]
    \centering
    \includegraphics[width=\linewidth]{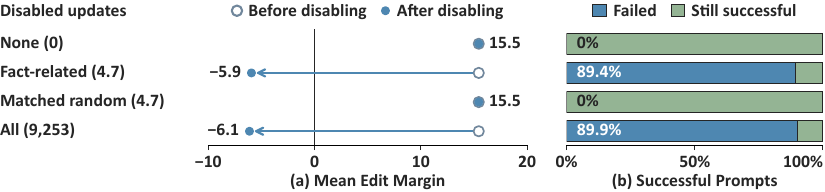}
    \caption{Disabling memory updates on CounterFact. (a) Mean edit margins before and after disabling. Positive values favor revised answers; negative values favor original answers. (b) Outcomes of the 5,850 prompts successful before disabling. Parentheses give the mean number of disabled updates per case.}
    \label{fig:memory_disabling_diagnostics}
\end{figure}

To test whether revised-fact recall depends on fact-related memory updates, we disable selected updates during inference. We use the model from Figure~\ref{fig:in_depth_analysis}(c), with 9,253 updated embeddings from 2,000 sequential CounterFact edits. Fact-related updates correspond to $n$-grams selected at the last subject token of each fact's original and four generated expressions. Four conditions disable 1)~no updates (\emph{None}); 2)~the evaluated fact's updates (\emph{Fact-related}); 3)~other updates matched in number and $n$-gram length (\emph{Matched random}); or 4)~all updates (\emph{All}). The random control (seed 0) excludes fact-related updates but does not match update norms or corpus frequencies. Original embeddings remain active, and updates are restored after each evaluation without further editing. Disabling a shared embedding's update removes the combined changes from all associated edits. Differences between All and separately recorded pre-edit outputs may reflect finite-precision variation. For each prompt $x$ and edit $e_i$, the edit margin measures revised-answer preference:
\begin{equation}
    \operatorname{margin}_i(x)
    =\operatorname{score}(x,o_i^\star)-\operatorname{score}(x,o_i),
    \label{eq:memory_disabling_margin}
\end{equation}
using length-normalized log-likelihood scores. Positive margins favor revised answers and negative margins original answers. We aggregate margins from edit prompts and held-out paraphrases within each case, then average cases equally. Of the 2,000 edit prompts and 4,000 held-out paraphrases, 5,850 succeed under None. The success-to-failure rate reports the percentage of these same prompts that no longer favor the revised answer after disabling. From Figure~\ref{fig:memory_disabling_diagnostics}, we observe that:

\begin{itemize}[leftmargin=1em,labelsep=0.4em,nosep]
    \item Fact-related disabling makes the model favor original over revised answers on average and turns 5,228 of the 5,850 successful prompts into failures. These updates therefore support revised-fact recall for most previously successful prompts. Since the original embeddings remain active, the reversal reflects removal of the learned updates, not removal of the underlying memory.
    \item Matched random disabling changes the mean margin by only $-0.0003$ and causes no successful prompt to fail. The contrast shows that recall depends on which updates are disabled, not simply how many. Fact-related updates remain available under this control, allowing the model to continue using them to recall revised facts.
    \item Disabling an average of 4.679 fact-related updates nearly matches disabling all 9,253 updates in both mean margin and success-to-failure rate. This shows that recalling revised facts depends primarily on a small set of memory updates, activated when the input contains the corresponding fact-related $n$-grams.
\end{itemize}

\begin{figure}[t]
    \centering
    \includegraphics[width=\linewidth]{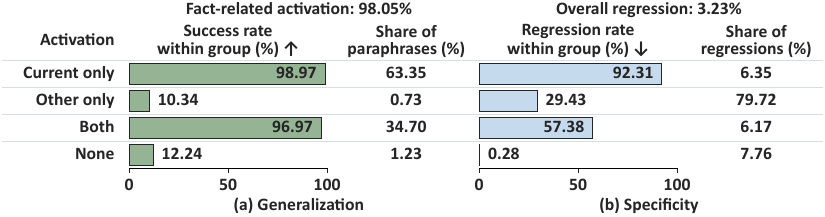}
    \caption{Memory activation and editing outcomes with all updates enabled. Groups indicate activation of embeddings updated for the current edit only, other edits only, both, or neither. (a) Within-group paraphrase success rates and each group's share of all paraphrases. (b) Success-to-failure rates among previously successful neighborhood queries in each group and each group's share of all regressions.}
    \label{fig:memory_activation_outcomes}
\end{figure}

\subsubsection{Memory Activation and Editing Outcomes}
\label{app:trigger_attribution}

To relate memory activation to generalization and preservation, we evaluate 4,000 held-out paraphrases and 20,000 neighborhood prompts after completing 2,000 sequential CounterFact edits, with all updates enabled. The \emph{current edit} is the edit paired with each prompt in CounterFact; neighborhood prompts ask about unrelated facts. Using the fact-related $n$-gram sets from Appendix~\ref{app:memory_disabling}, we scan all prompt positions before answer generation and group prompts by the updated embeddings they activate: 1)~\emph{Current only}, those associated only with the current edit; 2)~\emph{Other only}, only with other edits; 3)~\emph{Both}, with both; and 4)~\emph{None}, no updated embeddings. An embedding shared across edits counts toward each associated edit. The sets cover all 9,253 updated $n$-grams, and including teacher-forced answer prefixes leaves the groups unchanged.

\paragraph{Generalization.}
A held-out paraphrase succeeds when the revised answer scores above the original answer under length-normalized log-likelihood. The held-out paraphrases do not overlap with generated expressions under exact or normalized matching. Figure~\ref{fig:memory_activation_outcomes}(a) shows the success rate within each group and the group's share of all 4,000 paraphrases. We observe that:

\begin{itemize}[leftmargin=1em,labelsep=0.4em,nosep]
    \item Current only and Both together cover 98.05\% of held-out paraphrases, with success rates of 96.97--98.97\%. High success thus extends to most held-out paraphrases and is associated with activation of fact-related updates. Different wordings can share $n$-grams, so expressions not used during editing can still use embeddings updated for the same fact.
    \item Other only and None contain just 1.95\% of held-out paraphrases but account for 50.36\% of failures. This concentration of failures supports broader memory mapping across expressions, as tested in Section~\ref{sec:expression_coverage}. These paraphrases cannot directly use the updates for their fact because none of the corresponding $n$-grams are activated.
\end{itemize}

\paragraph{Specificity.}
On neighborhood prompts, success requires the original answer to score above the edit's revised answer. We compare the same prompts before and after editing and count initially successful prompts that become failures as \emph{regressions}. Figure~\ref{fig:memory_activation_outcomes}(b) shows the regression rate among initially successful prompts in each group and the group's share of all 567 regressions. We observe that:
\begin{itemize}[leftmargin=1em,labelsep=0.4em,nosep]
    \item Among the 17,534 initially successful prompts, 96.77\% remain successful and 3.23\% regress; the regression rate in None is only 0.28\%. Most previously correct predictions are thus retained, with especially strong preservation in None because these inputs do not directly use embedding updates. The 44 regressions in None may reflect finite-precision differences that reverse close answer rankings across inference runs.
    \item Other only accounts for 79.72\% of all regressions, whereas Current only accounts for 6.35\% despite its higher within-group regression rate. Overall, 92.24\% of regressions accompany activation of updated embeddings. Regressions are thus concentrated in inputs that activate updates for other edits. Different facts can share $n$-grams, so an update for one fact can also affect unrelated predictions. This supports using reuse-based regularization to penalize large updates to frequently reused embeddings.
\end{itemize}

\begin{figure}[t]
    \centering
    \includegraphics[width=\linewidth]{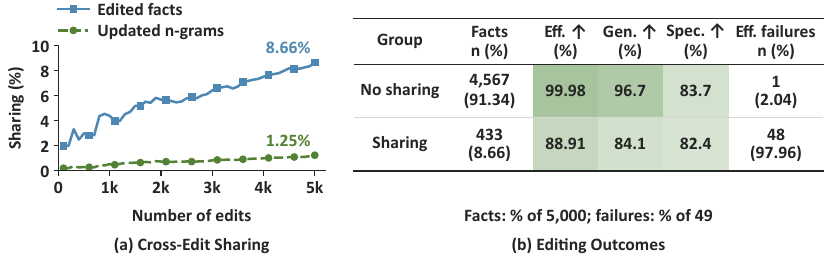}
    \caption{Cross-edit sharing on CounterFact. (a) Sharing rates among updated $n$-grams and edited facts. (b) Final editing results for facts with or without shared $n$-grams.}
    \label{fig:cross_edit_sharing}
\end{figure}

\subsubsection{Cross-Edit Sharing}
\label{app:cross_edit_sharing}

To examine how cross-edit sharing relates to editing outcomes, we analyze the 5,000-edit CounterFact run every 100 edits. We track $n$-grams selected at the last subject token of each fact's original and four generated expressions. A fact belongs to \emph{Sharing} if any selected $n$-gram is also selected for another fact, and to \emph{No sharing} otherwise. Figure~\ref{fig:cross_edit_sharing} reports sharing rates and groups the saved final results. We observe that:
\begin{itemize}[leftmargin=1em,labelsep=0.4em,nosep]
    \item Sharing increases overall but involves only 285 of 22,778 updated $n$-grams (1.25\%) and 433 facts (8.66\%) at 5,000 edits. Most embeddings are therefore selected for one fact; the higher fact-level rate reflects that a shared embedding serves multiple facts.
    \item No sharing covers 91.34\% of facts with 99.98\% Efficacy and 96.7\% Generalization, showing sustained editing success across expressions for most facts. Their selected embeddings are not shared with other edits, so different facts do not compete for updates to the same parameters.
    \item Sharing contains 48 of the 49 Efficacy failures (97.96\%), with lower Generalization but similar Specificity. Failures are thus concentrated in a small group of facts sharing updated embeddings, where different targets may require conflicting changes to the same parameters.
\end{itemize}

\end{document}